\documentclass{article} 
\usepackage{iclr2027_conference,times}

\usepackage{amsmath,amsfonts,bm}

\def\eqref#1{equation~\ref{#1}}

\def\1{\bm{1}}

\DeclareMathAlphabet{\mathsfit}{\encodingdefault}{\sfdefault}{m}{sl}
\SetMathAlphabet{\mathsfit}{bold}{\encodingdefault}{\sfdefault}{bx}{n}

\usepackage[T1]{fontenc}
\usepackage{amsmath,amssymb,amsthm}
\usepackage{booktabs}
\usepackage{xcolor,tabularx}
\newcolumntype{Y}{>{\raggedleft\arraybackslash}X}
\newcolumntype{Z}{>{\centering\arraybackslash}X}
\newcommand{\maintablestyle}{\small\renewcommand{\arraystretch}{1.06}\setlength{\tabcolsep}{3pt}}
\newcommand{\thcell}[1]{\shortstack[c]{#1}}
\newcommand{\thsingle}[1]{\raisebox{0.62em}[0pt][0pt]{#1}}
\usepackage{float}
\usepackage{graphicx}
\usepackage{etoolbox}
\usepackage{hyperref}
\hypersetup{colorlinks=false,pdfborder={0 0 1},
  linkbordercolor={1 0 0},citebordercolor={0 1 0},urlbordercolor={0 1 1}}
\usepackage{url}
\usepackage{natbib}
\usepackage{cleveref}

\newcommand{\tmem}{\tau_{\mathrm{mem}}}
\newcommand{\tgen}{\tau_{\mathrm{gen}}}
\newcommand{\Wpop}{W_{\mathrm{pop}}}
\newcommand{\Wemp}{W_{\mathrm{emp}}}

\newtheorem{proposition}{Proposition}

\newcommand{\newWindowGain}{$8.32\times$}
\newcommand{\newCopyReduction}{$75.9\%$}
\newcommand{\newAuditFID}{$75.56$}
\newcommand{\newSGDAuditFID}{$79.37$}
\newcommand{\newMuonCommonFID}{$81.3$}
\newcommand{\newQGDCommonFID}{$77.8$}

\title{Controlling Polar Exposure to Delay Memorization in Diffusion Models}

\author{Xuanchen Wang \\
School of Computer Science\\
The University of Sydney\\
Sydney, NSW, Australia  \\
\texttt{xwan0579@uni.sydney.edu.au} \\
\And
Heng Wang \\
School of Computer Science \\
The University of Sydney \\
Sydney, NSW, Australia \\
\texttt{heng.wang@sydney.edu.au} \\
\AND
Weidong Cai \\
School of Computer Science \\
The University of Sydney\\
Sydney, NSW, Australia \\
\texttt{tom.cai@sydney.edu.au}
}

\iclrfinalcopy 
\begin{document}

\maketitle
\lhead{}  
\renewcommand{\headrulewidth}{0pt}

\begin{abstract}
Diffusion models can reach useful sample quality before copying training examples, but fast optimization can compress this generalization window by accelerating sample-specific fitting. We investigate this effect through update geometry and propose Quality-Gated De-whitening (QGD), a controller that retains a fast polar-update prefix and progressively restores fixed-gain momentum. Our random-feature analysis separates covariance-controlled, curvature-equalized and amplitude-controlled memorization clocks. Under aligned spectral assumptions, it establishes a finite-exposure condition under which a fixed-gain tail recovers a delay proportional to dataset size. QGD implements this principle with a confirmed quality gate, a bounded decay envelope and causal copy feedback. Immediate switching is the conservative limit; gradual control balances delayed copying against continued quality improvement. We pair QGD with Copy-Budgeted Selection (CBS), which applies simultaneous binomial calibration to a frozen checkpoint family, followed by a fresh evaluation of the released checkpoint. On 2,000-image CIFAR-10 subsets, QGD preserves the polar baseline's quality-arrival time while expanding its useful interval by \newWindowGain{} and reducing common-checkpoint copying by \newCopyReduction{}. With identical calibration and independent quality evaluation, QGD achieves FID \newAuditFID{} versus \newSGDAuditFID{} for SGD with the same selector. Exposure-matched controls, independent detector audits and transfer to flow matching and dance generation support adaptive exposure control as a practical way to improve the quality--copying tradeoff.
\end{abstract}

\section{Introduction}
\label{sec:intro}
Diffusion models learn useful generative structure from finite training collections \citep{sohldickstein2015diffusion,song2019score,ho2020ddpm}, yet continued optimization can lead them to reproduce training examples \citep{carlini2023extracting,somepalli2023forgery}. Quality and copying emerge at different times, creating a generalization window in which a model produces useful samples before detectable memorization becomes prevalent \citep{bonnaire2025memorize,favero2025bigger}. This interval matters operationally: it provides room for further quality improvement and checkpoint selection. Reaching a quality target sooner is therefore one part of a successful training strategy; preserving useful choices after that point is another.

Prior work establishes that optimizer choice changes these clocks. \citet{bonnaire2025memorize} compare Adam and momentum SGD, observing faster learning and memorization with Adam while retaining dataset-size-dependent delay. Learning-rate effects also influence memorization in denoising score matching \citep{wu2025bigstep}. These findings motivate a more specific question: can update geometry change the mechanism governing the delay, and can that mechanism guide a controller that preserves fast initial learning? Answering it requires separating directional changes from update magnitude and measuring quality and copying along the same trajectories.

Our random-feature analysis identifies three clocks. Gradient flow retains covariance-dependent rates that slow fitting in small-eigenvalue directions. Exact curvature preconditioning equalizes those rates. Aligned polar updates instead make fluctuation fitting depend on coefficient amplitude, so distributing fluctuation energy across more directions can shorten its clock. This rank-dependent compression gives a concrete control objective: limit polar exposure after useful population structure has been learned, then restore the small-gradient delay of a fixed-gain tail. Controlled spectra and directional interventions test this mechanism separately from neural architecture changes.

We propose Quality-Gated De-whitening (QGD) to implement this objective. QGD begins with Muon, an existing optimizer that orthogonalizes matrix momentum \citep{jordan2024muon,liu2025muonscalable}. A predefined quality target, confirmed by a subsequent probe, determines when control starts. Thereafter, causal copy feedback reduces the polar coefficient inside a deterministic decay envelope while preserving momentum and a fixed gain. The envelope bounds total subsequent polar exposure; feedback adapts its allocation to the observed trajectory. Avoiding per-step renormalization allows small-gradient directions to remain slow.

Our recovery theorem gives a sufficient condition for a memorization delay proportional to dataset size: the fluctuation margin remaining at the gate must exceed subsequent polar exposure at the relevant scale. Immediate switching spends no additional polar exposure and provides the conservative endpoint of this analysis. QGD uses a gradual transition to retain quality improvement while delaying copying relative to continued polar training. We test this tradeoff against the same-gate hard switch and a pilot-fixed gradual schedule with matched exposure. Thus the theory motivates the controlled quantity, and the experiments test the practical value of its allocation.

Optimizer transitions toward SGD are established \citep{keskar2017swats,luo2019adabound}; MiMuon also adaptively selects Muon or momentum-SGD updates using a spectral criterion and practical norm proxy \citep{mimuon2026}. QGD connects a confirmed generative-quality target and causal copying probes to a finite-exposure condition. Appendix~\ref{sec:related} develops these relationships. This formulation makes a conservative switching baseline central to evaluation and tests feedback under varying copying dynamics rather than attributing every gain to adaptation.

\begin{figure}[t]
\centering
\includegraphics[width=\linewidth]{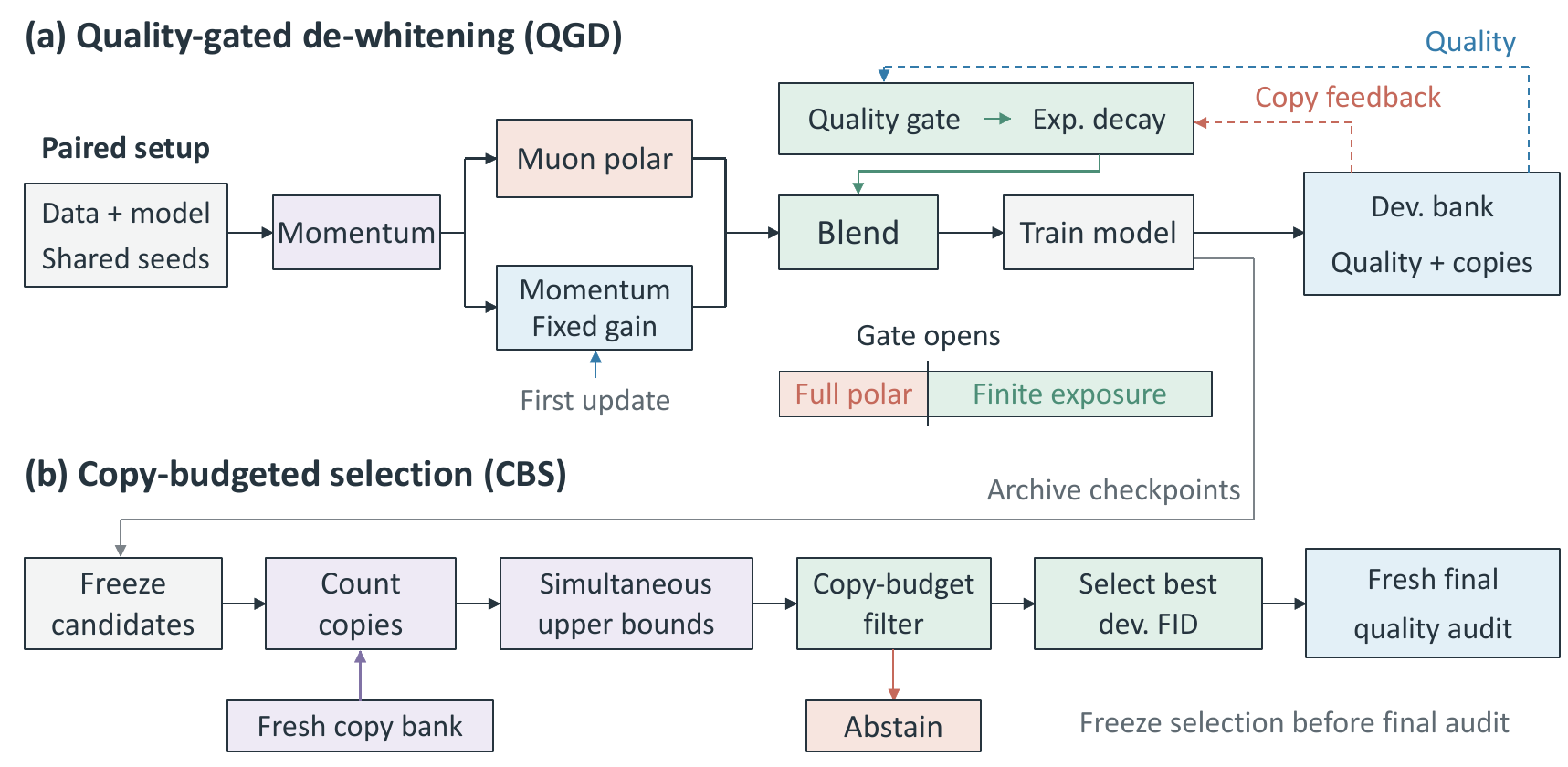}
\caption{\textbf{Quality-gated training and calibrated checkpoint release.} QGD preserves a fast polar prefix, then uses development copying feedback to reduce polar exposure within a fixed envelope. CBS evaluates a frozen checkpoint family with fresh calibration inputs. A separate final audit evaluates the selected checkpoint without changing the decision.}
\label{fig:overview}
\end{figure}

The resulting archive still requires a release decision. We pair QGD with Copy-Budgeted Selection (CBS), applying established exact binomial bounds and simultaneous calibration to a frozen checkpoint family \citep{clopper1934confidence,angelopoulos2025ltt}. CBS chooses by development quality among eligible checkpoints, or abstains, while bounding the probability of release with a detector-budget violation. Fresh calibration never guides training, and an additional independent bank evaluates final quality. Figure~\ref{fig:overview} shows these distinct roles. Comparisons with valid fixed-sequence testing and preselection expose the quality--availability--cost tradeoff of searching more candidates.

Across matched training conditions, QGD preserves Muon's quality arrival, widens its useful interval, and improves independently evaluated quality under a common copy budget. Conservative baselines retain lower copying in some settings. 

Our contributions can be summarized as follows:
\begin{itemize}
\item {}We develop an optimizer-geometric analysis of memorization in a random-feature denoising model, distinguishing covariance-controlled, curvature-equalized, and amplitude-controlled clocks. The analysis identifies rank-dependent compression of memorization delay under aligned polar updates and motivates controlling their cumulative exposure.

\item {}We propose Quality-Gated De-whitening (QGD), which combines a confirmed quality gate and causal copying feedback to transition from fast polar updates to fixed-gain momentum. We establish a finite-exposure condition under which the aligned model recovers a memorization delay proportional to dataset size.

\item {}We introduce Copy-Budgeted Selection (CBS), a checkpoint-selection procedure that combines simultaneous binomial calibration over a frozen candidate family with quality-based selection. CBS bounds the joint probability of release and detector-budget violation, while an independent final audit evaluates the selected checkpoint without reusing its selection measurements.

\item {}We evaluate QGD through matched-exposure controls, component ablations, detector audits, and transfer to flow matching and dance generation. On CIFAR-10 with $2{,}000$ training images, QGD preserves Muon's quality-arrival time, extends the observed generalization window to $8.32\times$ that of Muon, and reduces common-checkpoint copying by $75.9\%$. Under identical CBS, QGD achieves final-audit FID $75.56$, compared with $79.37$ for SGD.
\end{itemize}


\section{Related work}
\label{sec:related}

Diffusion and score-based models learn generative processes through denoising \citep{sohldickstein2015diffusion,song2019score,ho2020ddpm}, with quality and cost shaped by objectives, noise schedules, architectures, guidance and samplers \citep{nichol2021improved,dhariwal2021diffusion,karras2022edm}. Our paired protocol fixes these choices within each optimizer comparison, using a common pixel-space UNet despite broader latent and transformer designs \citep{rombach2022latent,peebles2023dit}. Diffusion models can reproduce training images \citep{carlini2023extracting,somepalli2023forgery}, especially with small datasets, duplication and large models \citep{yoon2023generalize,gu2025memorization,kadkhodaie2024generalization}. We use the copy criterion of \citet{yoon2023generalize} and study how optimizer choice shifts the onset of detectable copying, complementing prior mitigation work \citep{somepalli2023mitigating,wen2024detecting,dutt2024memcontrol}.

\citet{bonnaire2025memorize} identified an early quality time $\tgen$ and later memorization time $\tmem$, with a window that widens with dataset size under gradient-based training. Related analyses study the same transition \citep{achilli2025manifold,pham2026diffusion}, while \citet{wang2026consistency} shows that much diffusion behavior across data splits has a linear origin in empirical Gaussian statistics. We use the same tractable regime but ask how this quality-to-copy window depends on optimization geometry.

Muon orthogonalizes matrix momentum through Newton--Schulz iterations \citep{jordan2024muon} and corresponds to spectral-norm steepest descent \citep{bernstein2024modular}; related preconditioners include K-FAC \citep{martens2015kfac}, Shampoo \citep{gupta2018shampoo}, SOAP \citep{vyas2025soap}, and diagonal Adam \citep{kingma2015adam,loshchilov2019adamw}. Recent work reports faster memorization under Muon than AdamW on small datasets and altered memorization--generalization transitions \citep{optconsistency2026,tveit2025muongrokking,mimuon2026}. Optimizer transitions such as SWATS, AdaBound and MiMuon adapt optimization over training \citep{keskar2017swats,luo2019adabound,mimuon2026}. QGD differs by using a quality-triggered continuous mixture tailored to diffusion: it preserves a fast prefix, limits polar exposure using copy feedback, and retains eigenvalue-dependent motion in the tail. Our finite-exposure analysis explains this mechanism; related diffusion work also studies early stopping and large-step effects \citep{favero2025bigger,wu2025bigstep}.

CBS integrates detector-based checkpoint selection with exact binomial confidence intervals \citep{clopper1934confidence} and development/calibration separation \citep{angelopoulos2025ltt}; Bonferroni correction controls selection across a frozen checkpoint family, while adaptive Learn-then-Test provides a sequential alternative \citep{zecchin2025adaptive}. More broadly, mode-dependent learning dynamics in deep linear networks \citep{saxe2014exact,advani2020highdimensional}, early stopping as regularization \citep{yao2007early,raskutti2014early}, random-feature models \citep{rahimi2007random}, and supervised memorization dynamics \citep{arpit2017closer} motivate viewing training time as part of the model. We connect these perspectives to diffusion and identify how whitening reshapes its useful stopping region.

\section{Optimizer geometry determines the memorization clock}
\label{sec:theory}
A random-feature denoiser $D_W(y)=W\phi(y)$ isolates update geometry with fixed features \citep{rahimi2007random}. For Gaussian data with additive Gaussian noise, least-squares readout learning has covariance $K_n$ and empirical optimum $\Wemp=\Wpop+\Delta$, decomposed into population optimum and empirical fluctuation. We study a spectral-separation regime with population energy at order-one covariance eigenvalues and fluctuation energy at smaller ones. Appendix~\ref{app:linear-model} specifies the model and clocks.

\paragraph{Population Structure and Empirical Detail.}
For feature covariance $H$, write $\|A\|_H^2=\operatorname{tr}(AHA^\top)$ and $\langle A,B\rangle_H=\operatorname{tr}(AHB^\top)$. The empirical objective $L_n(W)=\|W-\Wemp\|_{K_n}^2+\mathrm{const}$ makes covariance determine gradient fitting rates. With population covariance $K$, normalized readouts separate population learning from progress toward the empirical fluctuation:
\begin{equation}
\mathrm{gen}(t)=\frac{\|W_t-\Wpop\|_K^2}{\|\Wpop\|_K^2},
\qquad
\mathrm{mem}(t)=\frac{\langle W_t-\Wpop,\Delta\rangle_{K_n}}{\|\Delta\|_{K_n}^2}.
\label{eq:main-toy-metrics}
\end{equation}
With positive denominators, these readouts measure population prediction error and progress toward empirical fluctuation. In aligned disjoint channels, memorization is the covariance-weighted fitted fraction $m(t)$, equal to one at the empirical optimum. Gradient-flow bounds also control population/fluctuation cross terms outside this alignment.

For a toy trajectory, let $g_\star=\inf_{t\ge0}\mathrm{gen}(t)$. A cell qualifies if $g_\star\le\mathrm{gen}(0)/2$. Then $\tgen$ first reaches $1.1g_\star$, with the positive tolerance defined in Appendix~\ref{app:linear-model} and $\tmem$ reaches half the fluctuation projection; an undefined quality clock leaves the window ratio undefined. These clocks order structure and fluctuation learning; the neural protocol uses a shared sample-quality target (Section~\ref{sec:protocol}).

\paragraph{Three Clocks.}
Normalized gradient flow learns covariance mode $k$ as $(1-e^{-\lambda_k t})\Wemp u_k$. Its fluctuation-energy quantile bounds give $\tmem=\Theta(n)$ when the relevant eigenvalues scale as $n^{-1}$ and the normalized population/fluctuation cross energy remains bounded away from $1/4$ (Proposition~\ref{prop:gflow}). Exact curvature preconditioning instead yields $W_t=(1-e^{-t})\Wemp$ (Proposition~\ref{prop:newton}). In the disjoint-channel case, its half-fluctuation time is $\log 2$; its straight trajectory need not reach the quality gate.

Polar flow supplies a different route. With aligned covariance and target singular directions, it learns coefficient $a_j$ as $\min\{\eta t,a_j\}$. If the fluctuation singular values lie in $[s_{\min},s_{\max}]$, then:
\begin{equation}
\frac{s_{\min}}{2\eta}\le\tmem\le\frac{s_{\max}}{2\eta},\qquad
\tmem=\Theta\!\left(\frac{\sqrt{n/r}}{\eta}\right)
\quad\text{if }\|\Delta\|_F^2=\Theta(n),\ s_{\max}/s_{\min}=O(1),
\label{eq:three-clocks}
\end{equation}
where $r$ is fluctuation rank (Proposition~\ref{prop:aligned-polar}). The bounds depend on coefficient distance and rank rather than the positive covariance eigenvalues. Practical Muon includes momentum, approximate orthogonalization and learned features; Figure~\ref{fig:mechanism} summarizes the geometric distinction and random-feature evidence.

\paragraph{Random-Feature Evidence.}
Across six dataset sizes and two noise levels, gradient-flow memorization slopes are $1.00$/$1.05$, versus Muon's $0.20$/$0.18$ at $d=64$; the paired window-slope gaps are $-0.75$/$-0.82$, with $21.14\times$/$29.72\times$ narrowing at $n=512$ (Table~\ref{tab:toy}). At $d=16$, Muon's slopes rise to $0.60$/$0.65$ while gradient flow remains near one. The ordering survives $\tanh$ and noise-draw changes (Tables~\ref{tab:toy-tanh}--\ref{tab:m512}); exact preconditioning misses quality for $n\le64$ (Table~\ref{tab:levels}). Doubling $n$ approximately halves the fluctuation eigenvalue while preserving the population eigenvalue (Table~\ref{tab:spectrum}). Across draws, sizes and noise levels, inverse fluctuation eigenvalues correlate with memorization time at $0.99$ for gradient flow versus $0.14$ for Muon (Spearman; Table~\ref{tab:diagnostics}).

\paragraph{From a Timing Diagnosis to a Control Objective.}
The mechanism motivates a polar prefix followed by restoration of small-eigenvalue delay. Spreading fluctuation energy over more aligned directions shortens each coefficient's polar travel distance; the readout-dimension comparison and the joint spectral family in Appendix~\ref{app:exposure-diagnostic} test this dependence.

\section{Exposure-controlled training and calibrated release}
\label{sec:protocol}
\subsection{Separate development, calibration and evaluation}
Paired runs share the training subset, initialization, sample order and training noise. Matrix methods use the same non-matrix AdamW updates and the same development-probe schedule. The primary model is a $6.47$M-parameter UNet trained on class-balanced CIFAR-10 subsets of $2{,}000$/$10{,}000$ images for $200$k/$400$k steps. Five paired trajectories are evaluated at each size. The image sampler is deterministic DDIM \citep{song2021ddim}; Appendix~\ref{app:impl} specifies the model, sampler and optimizer settings.

Development probes run every $h=2{,}000$ steps. FID \citep{heusel2017fid,parmar2022cleanfid} uses $5{,}000$ outputs; a $1{,}000$-output subset guides copying feedback. Pixel detection flags an output when its nearest and second-nearest training-image distances satisfy $d_1<d_2/3$ \citep{yoon2023generalize,gu2025memorization}. Independent pilots fix the quality targets, $88.0$/$64.9$, before evaluation. The first quality crossing defines $\tgen$ after next-checkpoint confirmation; the controller changes only subsequent updates. A confirmed copy crossing before or tied with quality confirmation makes the run inadmissible.

The first development-copy crossing of $5\%$, confirmed at the next probe, defines $\tmem$. For horizon $H$, the observed useful interval is $\Delta t_H=\max(0,\min(\tmem,H)-\tgen)$; an unreached copy crossing remains censored. Independent banks evaluate common-checkpoint quality and copying without affecting these clocks or control. After training, CBS calibrates a frozen checkpoint family with fresh inputs. A further independent bank evaluates each selected checkpoint's final quality; it never changes the selected identity. These roles separate an actionable controller, a valid release decision and an independent final assessment. Appendix~\ref{app:audit} defines the banks and fixed detector rules.

\subsection{Our training method: quality-gated de-whitening}
\label{sec:qgd}
QGD keeps the full Muon update until the predeclared quality target is confirmed at $t_s$. For matrix layer $\ell$, let $M_{\ell,k}$ be its momentum direction and $U^{\rm Muon}_{\ell,k}$ its practical Muon direction. Calibrate $c_\ell=\mathrm{RMS}(U^{\rm Muon}_{\ell,0})/\mathrm{RMS}(M_{\ell,0})$ at the first update and hold it fixed; use the positive fallback $c_\ell=1$ if either RMS is at most $\epsilon$. The mixed direction is:
\begin{equation}
U_{\ell,k}=\alpha_k U^{\rm Muon}_{\ell,k}+(1-\alpha_k)c_\ell M_{\ell,k}.
\label{eq:qgd-update}
\end{equation}
The mixture has no subsequent RMS normalization, which could divide out the small-gradient delay restored by the fixed-gain tail. Before confirmation, $\alpha=1$; a confirmed development-copy budget crossing before or tied with quality confirmation makes the run inadmissible.

At $t_j=t_s+jh$, use the available development copy fraction $\widehat C_j$ to set the coefficient for the \emph{following} block, starting from $\alpha_0=1$:
\begin{equation}
\alpha_{j+1}=\min\!\left\{e^{-(j+1)h/T},\ 
\left[\alpha_j-\kappa\left[\frac{\widehat C_j-c_{\rm warn}}{c_{\rm budget}-c_{\rm warn}}\right]_+\right]_+\right\},
\qquad [x]_+=\max(x,0).
\label{eq:qgd-recursion}
\end{equation}
The default configuration uses $T=8{,}000$, $\kappa=0.5$, $c_{\rm warn}=2.5\%$ and $c_{\rm budget}=5\%$; sensitivity is evaluated separately. The coefficient decreases causally, and its exponential envelope bounds total subsequent polar exposure:
\begin{equation}
A_\infty=\int_{t_s}^{\infty}\alpha(t)\,dt
=h\sum_{j\ge0}\alpha_{j+1}\le\frac{h}{e^{h/T}-1}=7{,}041.6\ \text{steps}.
\label{eq:qgd-exposure}
\end{equation}
Copy feedback can shorten this exposure without compromising its bound.

\begin{proposition}[Finite polar exposure restores a spectral delay]
\label{prop:qgd}
In the aligned disjoint-channel model of Proposition~\ref{prop:aligned-polar}, let each fluctuation channel have target amplitude $s$ and covariance eigenvalue $\mu$. Run polar flow at speed $\eta$ until $t_s$, then mix it with fixed-gain gradient flow of gain $\gamma>0$ as in \Cref{eq:qgd-flow}, with measurable $\alpha(t)\in[0,1]$ and $A_\infty<\infty$. Set $m_s=\eta t_s/s$ and $\nu=\eta/s$. If $m_s<\theta<1$, $\nu\ge\gamma\mu$, and $\varepsilon_\theta=\theta-m_s-\nu A_\infty>0$, the time to fluctuation level $\theta$ satisfies:
\begin{equation}
t_s+\frac{\varepsilon_\theta}{\gamma\mu}\le\tau_\theta
\le t_s+\frac{1}{\gamma\mu}\log\frac{1-m_s}{1-\theta}.
\label{eq:qgd-clock}
\end{equation}
For fixed $\theta\in(0,1)$, $\mu=\Theta(n^{-1})$, $\gamma=\Theta(1)$, $t_s=O(1)$ and uniformly positive $\varepsilon_\theta$, this gives $\tau_\theta=\Theta(n)$.
\end{proposition}
Bounding $\dot m=\alpha\nu+(1-\alpha)\gamma\mu(1-m)$ proves the result (Appendix~\ref{app:control-proofs}); Proposition~\ref{prop:qgd-heterogeneous} extends it to unequal amplitudes with $c/n\le\mu_j\le C/n$. A positive polar floor retains amplitude-controlled arrival, whereas finite exposure with the stated margin gives a sufficient pathwise guarantee. The exact result preserves the common prefix and population-channel progress; full neural quality is evaluated separately.

\paragraph{Delay Restoration and Quality Utility.}
The upper bound in \Cref{eq:qgd-clock} is exactly the same-gate hard-switch clock: under the aligned assumptions, retaining polar updates reaches the fluctuation threshold no later than hard switching (Appendix~\ref{app:exposure-diagnostic}). The theorem therefore establishes delay restoration; quality determines whether a gradual transition is useful. For a desired residual margin $\varepsilon_0>0$, define $B_\star=(\theta-m_s-\varepsilon_0)/\nu>0$. Choosing:
\begin{equation}
T\le \frac{h}{\log(1+h/B_\star)}
\label{eq:exposure-design-rule}
\end{equation}
makes the envelope sufficient for $\varepsilon_\theta\ge\varepsilon_0$. A constructed aligned diagnostic holds gain constant and measures every term: positive-margin exposure retains a linear delay, whereas excess exposure and a fixed polar floor compress it (Figure~\ref{fig:exposure-diagnostic}). Gradual exposure improves early population risk, while the hard switch achieves the latest threshold and lower eventual risk. In neural training, the exposure cap is observable; the theorem's fluctuation margin is not. Independent evaluation tests practical quality and copying.

\subsection{Our selection rule: copy-budgeted selection}
\label{sec:cbs}
Development probes guide training and can be reused adaptively. CBS uses a separate calibration bank after freezing $1\le M\le100$ candidate checkpoints, the detector, and the sampler. For each candidate $k$, count $S_k$ flagged outputs among $N=5{,}000$ new iid inputs, including latent noise and any conditioning drawn from the fixed generation distribution, independently of the development history. Let $p_k$ denote its underlying detector-flag probability. With $0<\delta<1$ and $a=\delta/M$, the one-sided Clopper--Pearson upper limit \citep{clopper1934confidence} is:
\begin{equation}
U_k=\begin{cases}
\mathrm{Beta}^{-1}(1-a;S_k+1,N-S_k),&S_k<N,\\
1,&S_k=N.
\end{cases}
\label{eq:cbs-bound}
\end{equation}
Among checkpoints meeting the fixed development-quality target and $U_k\le c_{\rm budget}$, choose the lowest development FID, breaking ties by earlier step; abstain if the set is empty. Exact binomial inversion and simultaneous testing supply the established statistical ingredients \citep{angelopoulos2025ltt}.

\begin{proposition}[Detected-copy control for a released checkpoint]
\label{prop:cbs}
Conditional on the training and development history that fixes the candidates and detector, CBS satisfies:
\begin{equation}
\Pr\{\widehat k\ne\varnothing\ \text{and}\ p_{\widehat k}>c_{\rm budget}\}\le\delta.
\label{eq:cbs-guarantee}
\end{equation}
\end{proposition}
A union bound on $\Pr\{p_k>U_k\}\le\delta/M$ proves the result (Appendix~\ref{app:control-proofs}). Fresh inputs may be shared across candidates; their independence is unnecessary. The copy budget and risk target govern different quantities: a $5\%$ budget concerns the selected checkpoint's detector-flag probability, while $\delta=0.05$ bounds calibration's probability of releasing a checkpoint above that budget. The guarantee concerns joint release and budget violation for the declared detector and generation distribution, excluding undetected copying and population FID. The frozen family includes every configuration searched using calibration. After a checkpoint is selected, a further fresh bank evaluates quality and copying without revising the decision.

\begin{table}[t]
\centering\maintablestyle
\caption{\textbf{Training-method comparison on CIFAR-10.} Times are thousands of updates. Mean $\pm$ sample SD over five paired trajectories. FID and copy fractions use independent audits at the stated checkpoint; clocks use confirmed development probes. NR denotes no confirmed copy crossing by the observation horizon $H=200$k/400k. Observed windows end at $\min(t_{\rm copy},H)$.}
\label{tab:main}
\begin{tabularx}{\linewidth}{@{}lYYYYY@{}}
\toprule
\thsingle{Method} & \thcell{Quality\\arrival $\downarrow$} & \thcell{Copy\\onset $\uparrow$} & \thcell{Window\\$\Delta t_H\uparrow$} & \thsingle{FID $\downarrow$} & \thcell{Copy (\%)\\$\downarrow$} \\
\midrule
\multicolumn{6}{l}{CIFAR-10, $n=2,000$; audit checkpoint: 60k} \\
SGD & $16.0\!\pm\!1.4$ & NR & $184.0\!\pm\!1.4$ & $85.4\!\pm\!0.3$ & $1.26\!\pm\!0.11$ \\
AdamW & $14.0\!\pm\!1.4$ & $124.0\!\pm\!6.2$ & $110.0\!\pm\!5.3$ & $82.9\!\pm\!0.3$ & $2.51\!\pm\!0.10$ \\
Muon & $8.0\!\pm\!1.4$ & $28.0\!\pm\!1.4$ & $20.0\!\pm\!1.4$ & $81.3\!\pm\!0.3$ & $16.10\!\pm\!0.47$ \\
QGD (ours) & $8.0\!\pm\!1.4$ & $174.4\!\pm\!10.1$ & $166.4\!\pm\!11.3$ & $77.8\!\pm\!0.3$ & $3.88\!\pm\!0.31$ \\
\midrule
\addlinespace[3pt]
\multicolumn{6}{l}{CIFAR-10, $n=10,000$; audit checkpoint: 120k} \\
SGD & $18.0\!\pm\!1.4$ & NR & $382.0\!\pm\!1.4$ & $63.3\!\pm\!0.2$ & $0.78\!\pm\!0.14$ \\
AdamW & $14.0\!\pm\!1.4$ & NR & $386.0\!\pm\!1.4$ & $61.0\!\pm\!0.2$ & $1.32\!\pm\!0.11$ \\
Muon & $12.0\!\pm\!1.4$ & $96.0\!\pm\!4.5$ & $84.0\!\pm\!4.0$ & $61.5\!\pm\!0.2$ & $12.37\!\pm\!0.44$ \\
QGD (ours) & $12.0\!\pm\!1.4$ & $360.0\!\pm\!9.1$ & $348.0\!\pm\!8.5$ & $56.1\!\pm\!0.2$ & $2.99\!\pm\!0.21$ \\
\bottomrule
\end{tabularx}
\end{table}

\section{Experiments and Results}
\label{sec:experiments}
We compare matched training trajectories, controlled exposure schedules and checkpoint release. Five paired trajectories are the unit of uncertainty; calibration repetitions remain nested within each trajectory. Tables identify QGD and CBS as (ours).

\subsection{Quality and copying along matched trajectories}
\label{sec:main}
\begin{figure}[t]
\centering
\includegraphics[width=\linewidth]{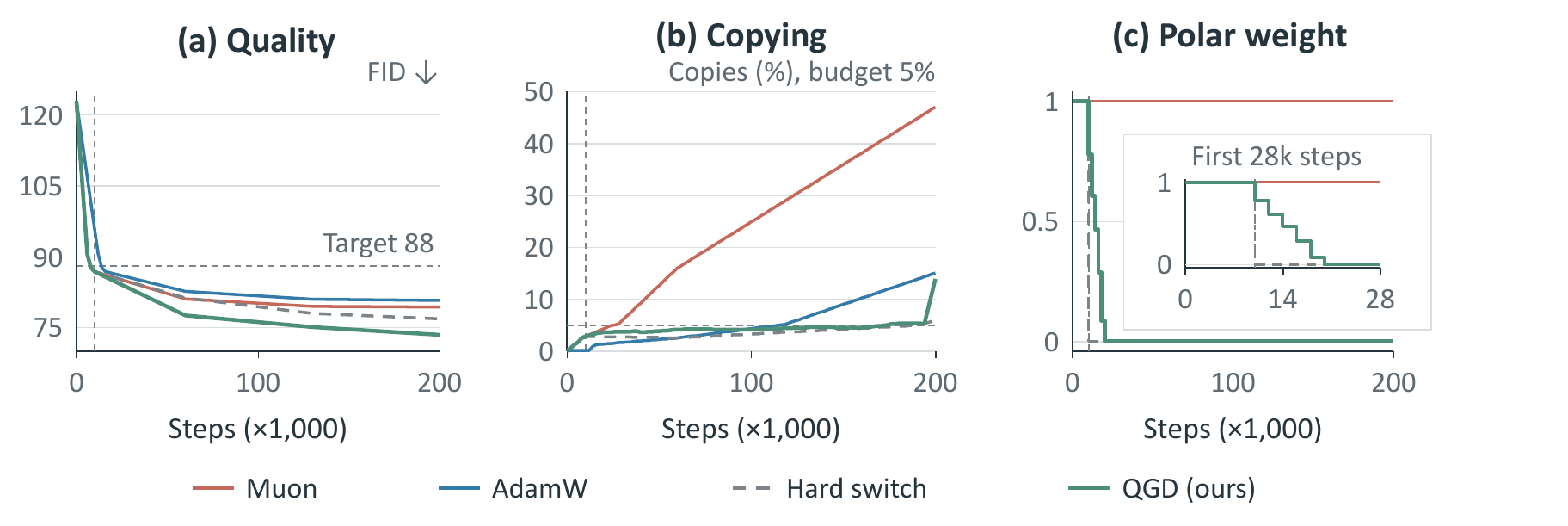}
\caption{\textbf{Quality, copying and exposure during training.} The primary CIFAR-10 trajectory (pair 0) compares QGD with continued Muon, AdamW and the same-gate hard switch. Development curves share a 0--200k time axis, with FID target 88 and copy budget 5\%. The 0--28k inset resolves the transition after quality confirmation at 10k. The hard switch is more conservative in copying; gradual QGD retains stronger quality. The coefficient for each block is fixed using the preceding development probe.}
\label{fig:cifar}
\end{figure}

QGD preserves Muon's quality-arrival time because their updates agree through confirmation. Its post-gate behavior produces the useful difference: at $n=2{,}000$, the mean interval expands by \newWindowGain{}, while independent common-step FID improves from \newMuonCommonFID{} to \newQGDCommonFID{} and copying falls by \newCopyReduction{} (Table~\ref{tab:main}). The larger subset preserves the same ordering against continued Muon. SGD and AdamW retain longer low-copy periods in some settings; QGD reaches useful quality earlier than SGD and achieves stronger later quality. This comparison concerns a common training time, while Table~\ref{tab:optimizer-cbs} evaluates the complete archive under a shared selector. Figure~\ref{fig:cifar} shows why a window needs both measurements: the hard switch suppresses copying more strongly, whereas QGD supports further quality improvement while remaining within the declared region. Development quality remains eligible from confirmation through the reported copy onset. Appendix~\ref{app:paired} provides paired clocks and per-trajectory effects, retaining censoring instead of treating observation limits as exact crossings.

\subsection{Exposure recovery and controller components}
\label{sec:timing}\label{sec:predictions}\label{sec:qgd-results}
The controlled spectral experiment verifies the condition used by Proposition~\ref{prop:qgd}. The finite-mixture schedule has residual margin $0.165$ across all tested sizes with fixed gain. Its copy time rises from $20.85$ to $148.49$ as $n$ grows from $64$ to $512$, while continued polar flow reaches the threshold at time $4$. Hard switching delays it further, as the comparison principle predicts. The gradual mixture fits the remaining population component earlier; the hard switch can achieve lower eventual error with more time. Figure~\ref{fig:exposure-diagnostic} and Table~\ref{tab:exposure-diagnostic} report both effects, and Appendix~\ref{app:exposure-diagnostic} gives the exact construction and margin calculation. Table~\ref{tab:qgd} tests the practical allocation of exposure. Hard switching shares the confirmed gate, momentum and first-update gain, then sets the polar coefficient to zero. It provides a longer copy delay and lower copying than QGD, while QGD improves common-step quality. The matched open-loop schedule uses the same gate and a pilot-fixed exposure budget, without access to evaluation copying feedback. Its mean exposure is comparable to QGD's; differences therefore test when exposure is spent rather than simply comparing a long schedule with a short one. The paired trajectories vary copying dynamics, allowing feedback to adapt its decay and shutoff time. Removing the quality gate sacrifices the shared fast prefix. Removing feedback while retaining the default envelope permits additional polar exposure, and per-step normalization removes the small-gradient scale that the fixed-gain tail is designed to preserve. Per-layer norm matching provides a separate directional intervention (Tables~\ref{tab:interpolation} and~\ref{tab:ns}). Controller settings, probe cadence and a matched MiMuon pilot search are specified in Appendix~\ref{app:controls}.

\subsection{Valid selection under a copying budget}
\label{sec:cbs-results}
Table~\ref{tab:cbs} compares selectors on the same QGD archives. Empirical filtering searches observed copying without a simultaneous correction. Single-candidate and margin preselection freeze a candidate using development information, then certify it once. Fixed-sequence testing fixes its ordering before calibration and stops at the first failed test; CBS applies simultaneous bounds to the full family \citep{angelopoulos2025ltt}. All valid rules use the same joint error target. Appendix~\ref{app:selection} specifies their ordering, testing and abstention rules. The comparison separates quality, release availability and calibration cost. At the tighter budget, searching a family can retain useful candidates that single preselection misses; a valid ordered search is a competitive alternative to uniform Bonferroni allocation. At the looser budget, margin preselection can approach the quality of a full search with far fewer outputs. These outcomes make the selector choice depend on the available archive and budget. Adaptive allocation offers another route to reducing testing cost \citep{zecchin2025adaptive}; here, fixed sample counts make the statistical comparison directly reproducible.

\begin{table}[t]
\centering\maintablestyle
\caption{\textbf{Switching controls and component ablations.} CIFAR-10, $n=2{,}000$; mean $\pm$ SD over five paired trajectories. Exposure and window are in k steps; quality and copying use independent audits at 60k. The open-loop schedule is fixed by an independent pilot. The no-gate exposure starts at initialization; MiMuon has no common post-gate mixture exposure.}
\label{tab:qgd}
\begin{tabularx}{\linewidth}{@{}lYYYY@{}}
\toprule
\thsingle{Method} & \thcell{Polar\\exposure} & \thcell{Observed window\\$\Delta t_H\uparrow$} & \thsingle{FID$_{60}\downarrow$} & \thcell{Copy$_{60}$ (\%)\\$\downarrow$} \\
\midrule
MiMuon (norm proxy) & -- & $84.4\!\pm\!5.0$ & $80.6\!\pm\!0.3$ & $4.30\!\pm\!0.30$ \\
Same-gate hard switch & $0.00\!\pm\!0.00$ & $186.0\!\pm\!2.4$ & $81.5\!\pm\!0.3$ & $2.47\!\pm\!0.03$ \\
Pilot-fixed open loop & $4.26\!\pm\!0.00$ & $124.0\!\pm\!50.0$ & $79.0\!\pm\!1.0$ & $4.56\!\pm\!0.60$ \\
QGD w/o quality gate & $7.03\!\pm\!0.00$ & $160.8\!\pm\!10.4$ & $82.9\!\pm\!0.3$ & $2.95\!\pm\!0.20$ \\
QGD w/o feedback & $7.04\!\pm\!0.00$ & $70.0\!\pm\!7.1$ & $79.2\!\pm\!0.3$ & $4.77\!\pm\!0.21$ \\
QGD + per-step RMS & $4.51\!\pm\!0.51$ & $30.0\!\pm\!2.4$ & $80.8\!\pm\!0.3$ & $13.51\!\pm\!0.16$ \\
QGD (ours) & $4.21\!\pm\!0.64$ & $166.4\!\pm\!11.3$ & $77.8\!\pm\!0.3$ & $3.88\!\pm\!0.31$ \\
\bottomrule
\end{tabularx}
\end{table}

\begin{table}[t]
\centering\maintablestyle
\caption{\textbf{Checkpoint-selection rules under copy budgets.} Five QGD archives, 100 independent calibration banks each; $\delta=0.05$. Final FID and fresh audit copying condition on release; FID is summarized over defined trajectory means. At 5\%, preselection never releases on one trajectory. Joint errors use the retained underlying detector rates. Calibration outputs exclude the common development archive and final audit.}
\label{tab:cbs}
\begin{tabularx}{\linewidth}{@{}lYYYYY@{}}
\toprule
\thsingle{Selection rule} & \thcell{Final FID\\$\downarrow$} & \thcell{Copy (\%)\\$\downarrow$} & \thcell{Release\\$\uparrow$} & \thcell{Joint error\\$\downarrow$} & \thcell{Calib. outputs\\(k) $\downarrow$} \\
\midrule
\multicolumn{6}{l}{Copy budget: 5\%} \\
Empirical filter & $73.97\!\pm\!0.30$ & 5.15 & 500/500 & 459/500 & 500.0 \\
Preselect + CP & $74.09\!\pm\!0.43$ & 5.11 & 20/500 & 3/500 & 5.0 \\
Margin preselect + CP & $77.89\!\pm\!2.28$ & 4.15 & 463/500 & 0/500 & 5.0 \\
Fixed sequence & $77.86\!\pm\!0.64$ & 4.08 & 500/500 & 0/500 & 152.6 \\
CBS (ours) & $75.56\!\pm\!0.51$ & 4.37 & 500/500 & 1/500 & 500.0 \\
\midrule
\addlinespace[3pt]
\multicolumn{6}{l}{Copy budget: 10\%} \\
Empirical filter & $73.85\!\pm\!0.30$ & 8.45 & 500/500 & 6/500 & 500.0 \\
Preselect + CP & $73.85\!\pm\!0.30$ & 8.38 & 421/500 & 0/500 & 5.0 \\
Margin preselect + CP & $73.88\!\pm\!0.32$ & 5.70 & 500/500 & 0/500 & 5.0 \\
Fixed sequence & $73.86\!\pm\!0.31$ & 7.83 & 500/500 & 0/500 & 480.2 \\
CBS (ours) & $73.86\!\pm\!0.31$ & 7.46 & 500/500 & 0/500 & 500.0 \\
\bottomrule
\end{tabularx}
\end{table}

\subsection{Independent quality of released checkpoints}
\label{sec:optimizer-cbs}
Holding CBS fixed compares the training methods under one release policy (Table~\ref{tab:optimizer-cbs}). QGD obtains final-audit FID \newAuditFID{}, compared with \newSGDAuditFID{} for conservative SGD, which can exploit its full low-copy archive. The independent audit preserves the quality advantage rather than reusing the development score that selected the checkpoint. The hard-switch pipeline also meets the budget but has higher final FID; QGD improves the quality available for release. The selected step identifies the released weights; it does not represent the total computation needed to produce and calibrate the archive. FID uncertainty is computed across five trajectory means after averaging their calibration repetitions. Release and violation frequencies count all decisions, including abstentions. Additional comparisons vary the candidate-family size, and Appendix~\ref{app:cost} reports archive construction, generation probes and calibration separately. This accounting makes the expense of adaptive control and full-family search visible alongside their quality gains.

\begin{table}[t]
\centering\maintablestyle
\caption{\textbf{Calibrated selection across training strategies.} CIFAR-10, $n=2{,}000$; 5\% copy budget and $\delta=0.05$. All methods use a 200k archive and identical CBS allocation. Final FID uses a fresh 50k-output bank; copying uses a 10k subset. Mean $\pm$ SD is across five trajectories after averaging 100 calibration decisions each. All 500 decisions per row release.}
\label{tab:optimizer-cbs}
\begin{tabularx}{\linewidth}{@{}lYYYY@{}}
\toprule
\thsingle{Training + selection} & \thcell{Selected\\step (k)} & \thcell{Develop.\\FID $\downarrow$} & \thcell{Final FID\\$\downarrow$} & \thcell{Copy (\%)\\$\downarrow$} \\
\midrule
SGD + CBS & 200.0 & 79.00 & $79.37\!\pm\!0.31$ & 2.67 \\
AdamW + CBS & 103.8 & 81.58 & $81.94\!\pm\!0.29$ & 4.24 \\
Muon + CBS & 18.5 & 85.81 & $86.14\!\pm\!0.12$ & 3.99 \\
MiMuon (norm proxy) + CBS & 64.2 & 80.40 & $80.68\!\pm\!0.40$ & 4.25 \\
Same-gate hard switch + CBS & 160.0 & 77.46 & $77.82\!\pm\!0.27$ & 4.30 \\
Pilot-fixed open loop + CBS & 77.8 & 79.00 & $79.30\!\pm\!3.08$ & 4.37 \\
QGD + CBS (ours) & 128.6 & 75.18 & $75.56\!\pm\!0.51$ & 4.37 \\
\bottomrule
\end{tabularx}
\end{table}

\subsection{Independent audits and transfer}
\label{sec:modality}\label{sec:robustness}
Independent detector audits evaluate the pixel-controlled QGD trajectory using fresh inputs and a fixed SSCD detector \citep{pizzi2022sscd}. Table~\ref{tab:detector-audit} reports both fixed detectors; Table~\ref{tab:sensitivity} varies the controller and probe settings. These tests distinguish the behavior of an adaptive controller from the AdamW--Muon ordering established by the background geometry experiments. Table~\ref{tab:transfer-main} evaluates flow matching \citep{lipman2023flow} and music-conditioned AIST++ motion \citep{li2021aist}. Image quality uses final-audit FID. Motion evaluation separates beat synchrony, kinetic-feature distribution quality and diversity, so improved alignment cannot conceal repetitive motion. The common quality target and detector are fixed within each task, with data splits formed before chunking. QGD improves the joint quality--copying tradeoff against continued polar training; conservative alternatives retain lower copying in some settings. Appendix~\ref{app:transfer} defines the task-specific metrics and copying rules.

\begin{table}[t]
\centering\maintablestyle
\caption{\textbf{Transfer across generation tasks.} Mean $\pm$ SD over five paired trajectories. Flow times are k updates; quality is audited at the 60k checkpoint. Dance quality at 120k separates beat synchrony, kinetic-feature FID and diversity relative to held-out motions; one is the diversity target. Copying uses 10k fresh audit outputs.}
\label{tab:transfer-main}
\begin{tabularx}{\linewidth}{@{}lYYYY@{}}
\toprule
\multicolumn{5}{l}{Flow matching, $n=2{,}000$} \\
\thsingle{Method} & \thsingle{FID $\downarrow$} & \thcell{Quality\\arrival $\downarrow$} & \thcell{Window\\$\Delta t_H\uparrow$} & \thcell{Copy (\%)\\$\downarrow$} \\
\midrule
AdamW & $84.5\!\pm\!0.3$ & $10.0\!\pm\!1.4$ & $106.0\!\pm\!2.8$ & $2.08\!\pm\!0.22$ \\
Muon & $86.2\!\pm\!0.3$ & $8.0\!\pm\!1.4$ & $18.0\!\pm\!2.8$ & $14.61\!\pm\!0.07$ \\
Same-gate hard switch & $82.8\!\pm\!0.3$ & $8.0\!\pm\!1.4$ & $120.0\!\pm\!2.8$ & $2.48\!\pm\!0.15$ \\
QGD (ours) & $79.8\!\pm\!0.3$ & $8.0\!\pm\!1.4$ & $100.0\!\pm\!2.8$ & $4.14\!\pm\!0.25$ \\
\midrule
\addlinespace[3pt]
\multicolumn{5}{l}{AIST++ dance generation} \\
\thsingle{Method} & \thcell{Beat\\$\uparrow$} & \thcell{Kinetic FID\\$\downarrow$} & \thcell{Diversity\\target: 1} & \thcell{Copy (\%)\\$\downarrow$} \\
\midrule
AdamW & $0.282\!\pm\!0.001$ & $31.7\!\pm\!0.4$ & $0.963\!\pm\!0.003$ & $1.94\!\pm\!0.13$ \\
Muon & $0.285\!\pm\!0.001$ & $34.6\!\pm\!0.4$ & $0.875\!\pm\!0.003$ & $11.06\!\pm\!0.57$ \\
Same-gate hard switch & $0.289\!\pm\!0.001$ & $30.2\!\pm\!0.4$ & $0.956\!\pm\!0.003$ & $2.14\!\pm\!0.22$ \\
QGD (ours) & $0.294\!\pm\!0.001$ & $27.9\!\pm\!0.4$ & $0.947\!\pm\!0.003$ & $4.09\!\pm\!0.29$ \\
\bottomrule
\end{tabularx}
\end{table}

\section{Discussion and conclusion}
\label{sec:discussion}\label{sec:conclusion}

Update geometry governs how quickly useful learning gives way to memorizing sample-specific detail. Finite polar exposure suffices to recover a spectral delay, linking optimizer geometry directly to the quality-to-copy interval. QGD turns this mechanism into quality-gated control: the hard switch suppresses copying most strongly, gradual QGD better preserves quality, and matched-exposure controls isolate the value of feedback rather than exposure reduction alone. With CBS fixed, an independent final bank measures the quality available for release across optimizers, while calibration shows when a larger checkpoint family justifies its additional cost. Together, these results suggest that optimizer choice, stopping time, and release calibration should be considered jointly rather than as separate design decisions.

\paragraph{Limitations.}
The theoretical delay guarantee assumes aligned, fixed features and idealized update dynamics. Extending the analysis to momentum, approximate orthogonalization, nonlinear feature learning, and representations that evolve throughout training would better capture modern diffusion systems. Our empirical study also focuses on a limited set of architectures, datasets, and optimizer families, so the magnitude of the observed effects may change at larger scales. CBS provides coverage only with respect to the specified copy detector and threshold (Appendix~\ref{app:selection}); it does not rule out memorization under alternative notions of similarity or extraction. Finally, probe cadence and checkpoint-family size determine the generation and calibration cost (Table~\ref{tab:cbs}), leaving a practical tradeoff between finer selection and evaluation expense.

\clearpage

\bibliography{iclr2027_conference}
\bibliographystyle{iclr2027_conference}

\clearpage
\appendix
\begin{center}
    {\fontsize{17}{22}\selectfont APPENDIX}
    \vspace{1.5em}
\end{center}

\section{Linear-model definitions and complete clock statements}
\label{app:linear-model}
\paragraph{Setup.}
Data $x\sim\mathcal{N}(0,\Sigma)$ in $\mathbb{R}^d$ are corrupted to $y=x+\sigma\varepsilon$ with $\varepsilon\sim\mathcal{N}(0,I)$. A denoiser $D_W(y)=W\phi(y)$ reads out $p$ fixed random features $\phi(y)=\mathrm{act}(Fy)$ through a matrix $W\in\mathbb{R}^{d\times p}$. Given $n$ training points, the empirical objective is:
\begin{equation}
L_n(W)=\frac{1}{n}\sum_{i=1}^{n}\mathbb{E}_{\varepsilon}\,\|W\phi(x_i+\sigma\varepsilon)-x_i\|^2
=\operatorname{tr}\!\big[(W-\Wemp)K_n(W-\Wemp)^{\!\top}\big]+\mathrm{const},
\label{eq:loss}
\end{equation}
where $K_n=\frac{1}{n}\sum_i\mathbb{E}_\varepsilon[\phi\phi^{\top}]$ is the feature covariance under the training measure and $\Wemp$ is the minimum-norm least-squares readout on the training set. The population readout $\Wpop$ minimizes the same objective under the data distribution, whose feature covariance is $K$. The two continuous flows below start at $W_0=0$ and converge to the minimum-norm $\Wemp$. Finite-step momentum and polar-update simulations require a separate stability and convergence check.

Assume positive population energy and $E_{\rm fl}=\operatorname{tr}[\Delta K_n\Delta^\top]>0$ whenever the corresponding normalized metric is used. Two scalar curves track a trajectory. Generalization is the prediction distance to the population readout on population inputs:
\begin{equation}
\mathrm{gen}(t)=\frac{\operatorname{tr}[(W_t-\Wpop)K(W_t-\Wpop)^{\!\top}]}{\operatorname{tr}[\Wpop K\Wpop^{\!\top}]},
\end{equation}
and memorization is the projection, on training inputs, of the current readout onto the part of the empirical readout that the population one does not contain:
\begin{equation}
\mathrm{mem}(t)=\frac{\operatorname{tr}[(W_t-\Wpop)K_n\,\Delta^{\!\top}]}{\operatorname{tr}[\Delta K_n\Delta^{\!\top}]},\qquad \Delta=\Wemp-\Wpop.
\end{equation}
Let $g_\star=\inf_{t\ge0}\mathrm{gen}(t)$ and fix $\xi=10^{-8}$ across cells. A cell qualifies when $g_\star\le\mathrm{gen}(0)/2$; otherwise $\tgen$ and $R$ are undefined. For qualifying cells, set $q_\star=\max(1.1g_\star,\xi\mathrm{gen}(0))$ and $\tgen=\inf\{t\ge0:\mathrm{gen}(t)\le q_\star\}$. The positive floor makes this level attainable even if a zero infimum is approached only asymptotically. The memorization clock is $\tmem=\inf\{t\ge0:\mathrm{mem}(t)\ge1/2\}$, with an empty set giving $+\infty$. Observed records retain censoring, which is excluded from exact-value slope fits. These are linear-model analogues of the quality and copy crossings in Section~\ref{sec:protocol}; a relative-to-own-minimum toy threshold is not the shared real-data threshold.

\paragraph{Where the Population and the Fluctuation Live.}
Write $(\lambda_k,u_k)$ for the eigenpairs of $K_n$ and $\delta_k=(\Wemp-\Wpop)u_k$. Population and fluctuation energies have weights proportional to $\lambda_k\|\Wpop u_k\|^2$ and $\lambda_k\|\delta_k\|^2$. We assume the spectral-separation regime: population energy lies on $O(1)$ eigenvalues and empirical fluctuation on eigenvalues that shrink with $n$. The following proposition quantifies its effect through fluctuation-energy quantiles. The spectral diagnostics in Appendix~\ref{app:toy-extra} describe this separation over a finite range; a joint dimensional regime or uniform spectral assumptions are required for an asymptotic statement.

\paragraph{Gradient Flow Separates the Two Timescales.}
Gradient flow on \Cref{eq:loss} moves mode $k$ at rate $\lambda_k$, so each part of the target is learned on the timescale of the eigenvalues that carry it (Figure~\ref{fig:mechanism}).

\begin{proposition}[Gradient flow]
\label{prop:gflow}
Under the normalized flow $\dot W=-\tfrac12\nabla L_n(W)$ from $W_0=0$, $W_tu_k=(1-e^{-\lambda_kt})\Wemp u_k$ for every $k$, and:
\[
\mathrm{mem}(t)=\sum_k\pi^{\mathrm{fl}}_k\,(1-e^{-\lambda_kt})-\rho(t),
\]
where $\rho(t)$ is a cross term between the population and the fluctuation that vanishes when population and fluctuation coefficients are orthogonal within each covariance eigenspace and is bounded by their normalized cross energy $\bar\rho$ for all $t$. For $q\in(0,1)$ let $\lambda^{\mathrm{fl}}_q$ be the eigenvalue at the $q$-quantile of the fluctuation energy counted from the top of the spectrum. If $\bar\rho<1/4$:
\[
\frac{1/4-\bar\rho}{\lambda^{\mathrm{fl}}_{1/4}}\;\le\;\tmem\;\le\;\frac{\log 3-\log(1-4\bar\rho)}{\lambda^{\mathrm{fl}}_{3/4}} .
\]
In particular, when the fluctuation quantiles scale as $\lambda^{\mathrm{fl}}_q\asymp n^{-1}$, the memorization time is $\Theta(n)$, provided $\bar\rho$ stays uniformly below $1/4$. Modes with eigenvalues of order one relax toward their empirical coefficients on an $O(1)$ timescale; closeness to $\Wpop$ additionally depends on fluctuation energy in those modes.
\end{proposition}

The proof is in Appendix~\ref{app:proofs}. Discrete gradient descent reproduces the flow to within $0.01$ in the memorization slopes, and Adam provides a diagonal-preconditioning comparison. Its ordering relative to other optimizers is empirical, not implied by this proposition.

\paragraph{Exact Curvature Preconditioning Removes Spectral Separation.}
An optimizer that preconditions by the curvature learns every mode at the same rate, and the separation disappears identically (Figure~\ref{fig:mechanism}).

\begin{proposition}[Exact preconditioning]
\label{prop:newton}
Under the normalized curvature flow $\dot W=-\tfrac12\nabla L_n(W)K_n^{+}$ from $W_0=0$, $W_t=(1-e^{-t})\Wemp$, so $\mathrm{mem}(t)=1-e^{-t}(1+c_n)$ with $c_n=-\mathrm{mem}(0)$, and, when $c_n>-1/2$, $\tmem=\log(2(1+c_n))$. If $c_n\le-1/2$, the threshold is already met at $t=0$. There is no direct eigenvalue-rate dependence, although $c_n$ can depend on the dataset. The straight path from $0$ to $\Wemp$ need not meet the generalization criterion; in that case $\tgen$ and $R$ are undefined.
\end{proposition}

This result motivates checking whether expressive practical preconditioners have a useful quality-before-copy interval on small datasets. It does not identify SOAP or Shampoo with the ideal flow or guarantee that either fails to generalize.

\paragraph{Orthogonalized Updates Retain a Coefficient-Distance Clock.}
Polar updates flatten singular values rather than directly inverting the covariance. A spectrally aligned limit makes their clock exact.
\begin{proposition}[Aligned polar flow]
\label{prop:aligned-polar}
Let $K_n v_j=\lambda_jv_j$, $\lambda_j>0$, and let $\{u_j\},\{v_j\}$ be orthonormal families. Write $\Wemp=\sum_j a_j u_jv_j^\top$, $a_j>0$, with $\Wpop$ retaining a subset of these channels and fluctuation $\Delta$ containing the complementary nonempty set $\mathcal F$. For compact polar flow $\dot W=-\eta\,\mathrm{polar}_0((W-\Wemp)K_n)$ from zero, with zero singular directions assigned zero updates:
\begin{equation}
 W_t=\sum_j\min\{\eta t,a_j\}\,u_jv_j^\top,
 \qquad \frac{s_{\min}}{2\eta}\le\tmem\le\frac{s_{\max}}{2\eta},
 \label{eq:aligned-polar-bounds}
\end{equation}
where $s_{\min},s_{\max}$ are the smallest and largest fluctuation singular values. Coefficient-arrival times and these bounds are independent of the positive covariance eigenvalues. If all fluctuation singular values equal $s$, then $\tmem=s/(2\eta)$ exactly.
\end{proposition}
If $\|\Delta\|_F^2=\Theta(n)$ and $s_{\max}/s_{\min}$ is bounded, the bound gives $\tmem=\Theta(\sqrt{n/r}/\eta)$ for fluctuation rank $r$. Thus a single aligned construction supports clocks of order $n$, $\sqrt{n/r}$ and $1$ for gradient, polar and curvature-preconditioned flow. Appendix~\ref{app:proofs} gives the proof and a finite-range rank crossover. Practical Muon adds momentum, approximate orthogonalization and evolving features; the experiments test the same geometric prediction in that setting.

\paragraph{Tempered Orthogonalization.}
We propose a scale-controlled intervention between normalized momentum and its polar factor. For nonzero momentum $M$, define $\widehat M=M/\|M\|_F$ and $\widehat P=\mathrm{polar}(M)/\|\mathrm{polar}(M)\|_F$, then normalize $(1-\lambda)\widehat M+\lambda\widehat P$ to a common per-layer RMS before applying the step. Zero matrices receive zero updates. This separates directional geometry from the changing norm of the original interpolation. The $\lambda=0$ endpoint is normalized momentum, not AdamW; AdamW remains a separate baseline. A second intervention varies the practical Newton--Schulz depth with the same RMS control. The interpolation turns update geometry into an experimental control. The results exhibit monotonic quality acceleration and window compression across both families. This is an empirical intervention result; the exact-flow propositions do not assert a general monotonicity theorem for the practical iteration.

\section{Proofs of the three optimizer clocks}
\label{app:proofs}

\paragraph{Proof of Proposition~\ref{prop:gflow}.}
From \Cref{eq:loss}, $\nabla L_n(W)=2(W-\Wemp)K_n$; using the normalized flow in the proposition, $\dot W=-(W-\Wemp)K_n$. In the eigenbasis of $K_n$ the columns $w_k(t)=W_tu_k$ decouple, $\dot w_k=-\lambda_k(w_k-w^{\mathrm{emp}}_k)$, giving $w_k(t)=(1-e^{-\lambda_kt})w^{\mathrm{emp}}_k$ from $w_k(0)=0$. Writing $w^{\mathrm{emp}}_k=w^{\mathrm{pop}}_k+\delta_k$:
\[
(W_t-\Wpop)u_k=\delta_k-e^{-\lambda_kt}\,(w^{\mathrm{pop}}_k+\delta_k),
\]
and therefore, with $E_{\mathrm{fl}}=\sum_j\lambda_j\|\delta_j\|^2$ and $\pi^{\mathrm{fl}}_k=\lambda_k\|\delta_k\|^2/E_{\mathrm{fl}}$,
\[
\mathrm{mem}(t)=\frac{1}{E_{\mathrm{fl}}}\sum_k\lambda_k\Big[\|\delta_k\|^2(1-e^{-\lambda_kt})-e^{-\lambda_kt}\langle w^{\mathrm{pop}}_k,\delta_k\rangle\Big]
=\sum_k\pi^{\mathrm{fl}}_k(1-e^{-\lambda_kt})-\rho(t),
\]
with $\rho(t)=E_{\mathrm{fl}}^{-1}\sum_k\lambda_ke^{-\lambda_kt}\langle w^{\mathrm{pop}}_k,\delta_k\rangle$, which is zero if $\langle w^{\mathrm{pop}}_k,\delta_k\rangle=0$ for all $k$ and satisfies $|\rho(t)|\le\bar\rho:=E_{\mathrm{fl}}^{-1}\sum_k\lambda_k|\langle w^{\mathrm{pop}}_k,\delta_k\rangle|$ for all $t\ge0$.

By the quantile definition, modes with $\lambda_k>\lambda^{\mathrm{fl}}_{1/4}$ carry fluctuation mass at most $1/4$ and modes with $\lambda_k\ge\lambda^{\mathrm{fl}}_{3/4}$ carry mass at least $3/4$. For the lower bound, use $1-e^{-s}\le\min(1,s)$ and split at $\lambda^{\mathrm{fl}}_{1/4}$:
\[
\mathrm{mem}(t)\le\sum_{\lambda_k>\lambda^{\mathrm{fl}}_{1/4}}\pi^{\mathrm{fl}}_k+t\sum_{\lambda_k\le\lambda^{\mathrm{fl}}_{1/4}}\pi^{\mathrm{fl}}_k\lambda_k+\bar\rho
\le\frac14+\lambda^{\mathrm{fl}}_{1/4}\,t+\bar\rho ,
\]
so $\mathrm{mem}(t)\ge1/2$ forces $t\ge(1/4-\bar\rho)/\lambda^{\mathrm{fl}}_{1/4}$. For the upper bound, keep only the modes with $\lambda_k\ge\lambda^{\mathrm{fl}}_{3/4}$, each of which has $1-e^{-\lambda_kt}\ge1-e^{-\lambda^{\mathrm{fl}}_{3/4}t}$:
\[
\mathrm{mem}(t)\ge\tfrac34\,\big(1-e^{-\lambda^{\mathrm{fl}}_{3/4}t}\big)-\bar\rho ,
\]
which reaches $1/2$ once $e^{-\lambda^{\mathrm{fl}}_{3/4}t}\le\frac13(1-4\bar\rho)$, that is for $t\ge\big(\log 3-\log(1-4\bar\rho)\big)/\lambda^{\mathrm{fl}}_{3/4}$. For modes with eigenvalues bounded away from zero, the response approaches its empirical coefficient at an $O(1)$ exponential rate. Translating that response into a bound on population prediction error additionally requires controlling the fluctuation component and its interaction with the population covariance $K$. \qed

\paragraph{Proof of Proposition~\ref{prop:newton}.}
With the preconditioner $K_n^{+}$, $\dot W=-(W-\Wemp)K_nK_n^{+}=-(W-\Wemp)$ on the range of $K_n$; the minimum-norm $\Wemp$ has its rows in that range and $W_0=0$ has no component off it, so $W_t=(1-e^{-t})\Wemp$. Substituting into the definition of $\mathrm{mem}$ gives:
\[
\begin{aligned}
\mathrm{mem}(t)&=\frac{\operatorname{tr}[(\Delta-e^{-t}\Wemp)K_n\Delta^{\!\top}]}{E_{\mathrm{fl}}}=1-e^{-t}(1+c_n),\\
c_n&=\frac{\operatorname{tr}[\Wpop K_n\Delta^{\!\top}]}{E_{\mathrm{fl}}}=-\mathrm{mem}(0).
\end{aligned}
\]
If $c_n>-1/2$, the first half-threshold crossing is $t=\log(2(1+c_n))$; otherwise it is already met at zero. The generalization curve is $\mathrm{gen}(t)=\|((1-s)\Wemp-\Wpop)K^{1/2}\|_F^2/\|\Wpop K^{1/2}\|_F^2$ with $s=e^{-t}$, a quadratic in $s$ whose minimizer and minimum value are determined by $\Wemp$, $\Wpop$, and $K$. When the minimum over $s\in[0,1]$ exceeds half of $\mathrm{gen}(0)$ the trajectory never enters the generalization phase under our definition. \qed

\paragraph{Proof of Proposition~\ref{prop:aligned-polar}.}
For the channel partition in the proposition, direct substitution gives:
\[
\mathrm{mem}(t)=\sum_{j\in\mathcal F}\omega_j\min\{\eta t/s_j,1\},
\qquad \omega_j=\frac{\lambda_js_j^2}{\sum_{\ell\in\mathcal F}\lambda_\ell s_\ell^2}.
\]

In the aligned subspace write $W_t=\sum_j b_j(t)u_jv_j^\top$.
The gradient direction is
$\sum_j\lambda_j(b_j-a_j)u_jv_j^\top$, whose compact polar factor is
$\sum_j\operatorname{sign}(b_j-a_j)u_jv_j^\top$ because
$\lambda_j>0$. Thus
$\dot b_j=\eta\operatorname{sign}(a_j-b_j)$, with zero velocity after
arrival, and $b_j(0)=0$. Its solution is
$b_j(t)=\min\{\eta t,a_j\}$, establishing
the coefficient-clipping formula and preservation of alignment.
Orthogonality of the two channel families eliminates the population cross
term in the memorization projection, giving
the weighted memory expression below. Before
$t=\theta s_{\min}/\eta$, every summand's clipped response is below
$\theta$. At $t=\theta s_{\max}/\eta$, every response is at least
$\theta$. Since the positive weights sum to one, these facts give
\Cref{eq:aligned-polar-bounds}. \qed

\paragraph{Rank Bound.}
Writing $r=|\mathcal F|$, suppose $c_E n\le\|\Delta\|_F^2\le C_E n$ and $s_{\max}/s_{\min}\le\kappa$, with constants independent of $n,r$. The RMS fluctuation singular value is $\|\Delta\|_F/\sqrt r$, and:
\[
\frac{\|\Delta\|_F}{\kappa\sqrt r}\le s_{\min}\le s_{\max}\le\frac{\kappa\|\Delta\|_F}{\sqrt r}.
\]
Consequently:
\[
\frac{\sqrt{c_E}}{2\kappa\eta}\sqrt{\frac nr}\le\tmem\le\frac{\kappa\sqrt{C_E}}{2\eta}\sqrt{\frac nr}.
\]
This $r$ is fluctuation rank; interpreting it as total update rank requires the two ranks to be comparable. With unequal $s_j$, the exact crossing still depends on the metric weights $\omega_j$, while the displayed bounds remain uniform in positive curvature.
\paragraph{An Explicit Three-Clock Comparison.}
If the fluctuation channels have a common covariance eigenvalue
$\lambda_{\mathrm{fl}}=c/n$ and common singular value
$s=a\sqrt{n/r}$, the same target gives:
\[
 \tau_{1/2}^{\mathrm{GF}}=\frac{n\log2}{\eta_{\mathrm{GF}}c},
 \qquad
 \tau_{1/2}^{\mathrm{polar}}=
       \frac{a}{2\eta_{\mathrm{polar}}}\sqrt{\frac nr},
 \qquad
 \tau_{1/2}^{\mathrm{prec}}=
       \frac{\log2}{\eta_{\mathrm{prec}}}.
\]
Here the comparison flows are respectively
$\dot W=-\eta_{\mathrm{GF}}(W-W_{\mathrm{emp}})K_n$ and
$\dot W=-\eta_{\mathrm{prec}}(W-W_{\mathrm{emp}})K_nK_n^+$.
Population channels with order-one target coefficients and order-one
covariance eigenvalues have order-one coefficient-learning clocks after
fixing the corresponding learning-rate normalization. Thus this
construction retains a coefficient-distance delay under polar flow while
removing the inverse-curvature delay. A statement about the full
population-risk threshold $\tau_{\mathrm{gen}}$ additionally requires
that threshold to remain comparable across $n$; coefficient arrival alone
does not establish that property.

\paragraph{A Finite-Range Rank Crossover.}
The aligned equal-singular-value family
$s(n,r)=a\sqrt{1+n/r}$ has the exact clock
$\tau_{1/2}=a\sqrt{1+n/r}/(2\eta)$. Its local log--log slope in $n$ is:
\[
 \frac{\partial\log\tau_{1/2}}{\partial\log n}
 =\frac12\frac{n}{n+r}.
\]
This explicit family has a nearly flat clock when $n\ll r$ and approaches
the square-root regime when $n\gg r$; increasing rank lowers the slope
over a fixed finite range of $n$. Its fluctuation energy is
$\|\Delta\|_F^2=a^2(n+r)$, so the order-$n$ energy assumption applies
in the regime $n\gtrsim r$. For the pure $\sqrt{n/r}$ family, changing a
fixed rank changes the prefactor, not the asymptotic exponent.

\paragraph{Why Tempering is an Empirical Intervention.}
For a general matrix gradient, singular vectors need not coincide with eigenvectors of $K_n$. The polar factor therefore does not define an independent constant-rate update in each covariance eigenmode. A convex combination of normalized momentum and polar directions does not imply monotonic crossing times, and finite Newton--Schulz iterations need not monotonically move every singular value toward one. The proposed comparison fixes the update RMS and reports the observed response rather than asserting a monotonicity theorem.

\section{Controller and checkpoint-selection proofs}
\label{app:control-proofs}
\paragraph{Mixed Flow and its Relation to the Practical Controller.}
The aligned-flow analysis uses exact compact polar directions and a fixed gradient gain:
\begin{equation}
 \dot W=-\alpha(t)\eta\,\mathrm{polar}_0((W-\Wemp)K_n)
 -(1-\alpha(t))\gamma(W-\Wemp)K_n,
 \label{eq:qgd-flow}
\end{equation}
It omits practical momentum and approximate Newton--Schulz orthogonalization. We assume $h,T>0$, $\kappa\ge0$, $0\le c_{\rm warn}<c_{\rm budget}<1$, and $0<\delta<1$. \Cref{eq:qgd-recursion} is nonincreasing and satisfies $\alpha_{j+1}\le e^{-(j+1)h/T}$. Summing this geometric series gives \Cref{eq:qgd-exposure}, independently of the accuracy of development copy estimates.

\paragraph{Proof of Proposition~\ref{prop:qgd}.}
Both directions preserve alignment. Before the threshold $\theta<1$, every fluctuation channel has coefficient $b(t)<s$ and follows:
\[
 \dot b(t)=\alpha(t)\eta+(1-\alpha(t))\gamma\mu(s-b(t)).
\]
The coefficients start from the common value $b(t_s)=\eta t_s$ and remain equal. The weighted memorization projection therefore equals $m(t)=b(t)/s$; the population cross term vanishes by the disjoint-channel assumption. It follows that:
\[
 \dot m=\alpha(t)\nu+(1-\alpha(t))\gamma\mu(1-m).
\]
Integrating $\dot m\le\alpha(t)\nu+\gamma\mu$ gives:
\[
 m(t)\le m_s+\nu A_\infty+\gamma\mu(t-t_s).
\]
Consequently reaching $\theta$ requires the lower bound in \Cref{eq:qgd-clock}. For the other direction, $\nu\ge\gamma\mu\ge\gamma\mu(1-m)$ implies:
\[
 \dot m\ge\gamma\mu(1-m),\qquad
 m(t)\ge1-(1-m_s)e^{-\gamma\mu(t-t_s)}.
\]
The right-hand side reaches $\theta$ at the claimed upper bound. This also establishes finite arrival. The bounds hold pathwise for every measurable coefficient schedule obeying the exposure constraint, including a schedule shortened by causal copy feedback. Under the stated uniform constants, both bounds are of order $n$. The always-polar claim follows from $m(t)=\min\{\eta t/s,1\}$.  \qed

\paragraph{Why a Fixed Polar Floor is Insufficient.}
If the post-gate coefficient has a fixed floor $\alpha(t)\ge a>0$, then $\dot m\ge a\eta/s$ up to the threshold, so:
\[
 \tau_\theta-t_s\le\frac{(\theta-m_s)s}{a\eta}.
\]
Such a floor retains an amplitude-controlled upper bound and cannot yield the stated linear-in-$n$ delay when $s=\Theta(\sqrt{n/r})$ and $\eta=\Theta(1)$. In contrast, the exponential envelope in \Cref{eq:qgd-recursion} bounds total post-gate exposure, regardless of whether the copy-feedback term activates.

\paragraph{Quality and Scale Qualifications.}
The QGD and full-polar trajectories coincide through $t_s$. Subsequently every population-channel coefficient continues monotonically toward its target, so its population-projection error does not increase. Full prediction error also includes fluctuation energy, and neural FID is outside this statement. Per-step normalization of the gradient branch can invalidate the spectral conclusion: with equal residual fluctuation singular values and already fitted population channels, normalized gradient and normalized polar updates have identical directions. Normalizing the final mixture, or matching the reference scale at a switch dominated only by fluctuation gradients, can remove the small-$\mu$ factor. These are distinct ablations rather than covered implementations.

\begin{proposition}[Heterogeneous fluctuation channels]
\label{prop:qgd-heterogeneous}
Retain the aligned disjoint-channel model and the mixed flow \Cref{eq:qgd-flow}, but permit fluctuation channel $j$ to have arbitrary $s_j>0$ and $\mu_j>0$. Let:
\[
 \omega_j=\frac{\mu_js_j^2}{\sum_{\ell\in\mathcal F}\mu_\ell s_\ell^2},\quad
 m_s=\sum_{j\in\mathcal F}\omega_j\min\{\eta t_s/s_j,1\},\quad
 \overline\nu=\sum_{j\in\mathcal F}\omega_j\eta/s_j.
\]
Write $\mu_{\min}=\min_{j\in\mathcal F}\mu_j$ and $\mu_{\max}=\max_{j\in\mathcal F}\mu_j$. If $m_s<\theta<1$, $\eta/s_j\ge\gamma\mu_j$ for every fluctuation channel, and $\varepsilon_\theta=\theta-m_s-\overline\nu A_\infty>0$, then:
\[
 t_s+\frac{\varepsilon_\theta}{\gamma\mu_{\max}}
 \le\tau_\theta\le
 t_s+\frac{1}{\gamma\mu_{\min}}\log\frac{1-m_s}{1-\theta}.
\]
Consequently, the linear-in-$n$ delay holds for a uniformly bounded fluctuation spectral band $c/n\le\mu_j\le C/n$, provided the other uniform conditions of Proposition~\ref{prop:qgd} hold.
\end{proposition}

\paragraph{Proof.}
Let $r_j(t)=b_j(t)/s_j\in[0,1]$. Orthogonality gives $m(t)=\sum_j\omega_jr_j(t)$. Until channel $j$ reaches its target:
\[
 \dot r_j=\alpha(t)\eta/s_j+(1-\alpha(t))\gamma\mu_j(1-r_j).
\]
After arrival, both directions are zero and $r_j=1$. Hence, at almost every time, including after individual channels saturate,
\[
 \gamma\mu_j(1-r_j)\le\dot r_j
 \le\alpha(t)\eta/s_j+\gamma\mu_j(1-r_j).
\]
The left inequality uses $\eta/s_j\ge\gamma\mu_j\ge\gamma\mu_j(1-r_j)$ before arrival and equality at saturation. Averaging the inequalities yields:
\[
 \gamma\mu_{\min}(1-m)\le\dot m
 \le\alpha(t)\overline\nu+\gamma\mu_{\max}(1-m)
 \le\alpha(t)\overline\nu+\gamma\mu_{\max}.
\]
Integrating the upper bound establishes the lower bound on the crossing time. Exponential comparison with the lower differential inequality establishes its upper bound. The result reduces to Proposition~\ref{prop:qgd} when all fluctuation channels have the same amplitude and curvature.  \qed

\paragraph{Proof of Proposition~\ref{prop:cbs}.}
Condition on the full history used to construct the family, including training data, optimizer randomness, reused development probes, candidate-selection decisions, and detector calibration. The family and detector are now fixed, while certification inputs are iid and independent of that history. Thus $S_k\sim\mathrm{Binomial}(N,p_k)$ for each fixed $k$. Exact one-sided binomial inversion gives $\Pr\{p_k>U_k\}\le\delta/M$. A union bound, which does not require independence across candidates, gives:
\[
 \Pr\{\exists k:p_k>U_k\}\le\delta.
\]
On its complement, every selected candidate has $p_{\widehat k}\le U_{\widehat k}\le c_{\rm budget}$. This proves the conditional statement; averaging over the history also gives the unconditional one.  \qed

\paragraph{Scope of the Certificate.}
\Cref{eq:cbs-guarantee} bounds the joint event of releasing a checkpoint whose detected-copy probability exceeds the budget. It is not a probability bound conditional on the event of release. It concerns the frozen detector, released weights, training reference set, sampler, and conditioning distribution, and does not certify undetected semantic copying or unseen deployment distributions. It supplies no population-FID guarantee. After certification results are revealed, new adaptively trained checkpoints require new independent certification samples or a separately justified sequential method.

\section{Hard-switch ordering and an exposure design rule}
\label{app:exposure-diagnostic}

\begin{proposition}[Same-gate hard switching maximizes the aligned delay]
\label{prop:hard-switch-ordering}
Retain the equal-channel assumptions of Proposition~\ref{prop:qgd}, including $\nu\ge\gamma\mu$, and compare trajectories starting from the same $m_s$ at $t_s$. For every measurable $\alpha(t)\in[0,1]$, the mixed-flow fluctuation fraction satisfies:
\begin{equation}
m_\alpha(t)\ge m_{\rm hard}(t)
=1-(1-m_s)e^{-\gamma\mu(t-t_s)},
\qquad
\tau_\theta^{\alpha}\le\tau_\theta^{\rm hard}
=t_s+\frac{1}{\gamma\mu}\log\frac{1-m_s}{1-\theta}.
\label{eq:hard-switch-ordering}
\end{equation}
The comparison holds up to threshold arrival and does not require a finite exposure integral. Under the positive-margin condition, Proposition~\ref{prop:qgd} bounds how much delay a finite-exposure mixture retains relative to this reference.
\end{proposition}

\paragraph{Proof.}
Before saturation:
\[
\dot m_\alpha
=\gamma\mu(1-m_\alpha)
+\alpha(t)\bigl[\nu-\gamma\mu(1-m_\alpha)\bigr]
\ge\gamma\mu(1-m_\alpha).
\]
The bracket is nonnegative because $m_\alpha\in[0,1]$ and $\nu\ge\gamma\mu$. Multiplication by the integrating factor $e^{\gamma\mu(t-t_s)}$, or scalar comparison from the shared initial value, yields \Cref{eq:hard-switch-ordering}. At saturation the fraction remains one, which also preserves the ordering.  \qed

\paragraph{Choosing an Admissible Envelope.}
Let $0<\varepsilon_0<\theta-m_s$ and $\nu>0$. The desired inequality
$\theta-m_s-\nu A_\infty\ge\varepsilon_0$ follows from
$A_\infty\le B_\star=(\theta-m_s-\varepsilon_0)/\nu$.
For the blockwise exponential envelope, Equation~\ref{eq:qgd-exposure} gives:
\[
\frac{h}{e^{h/T}-1}\le B_\star
\quad\Longleftrightarrow\quad
e^{h/T}\ge1+\frac{h}{B_\star}
\quad\Longleftrightarrow\quad
T\le\frac{h}{\log(1+h/B_\star)}.
\]
Copy feedback can reduce exposure below the envelope. This is a sufficient design rule when the aligned-model quantities are identified. Development detector counts do not identify $m_s$ or $\nu$ in a neural network, so the practical controller's measured exposure alone does not certify this model-specific margin. The heterogeneous result uses $\overline\nu$ in place of $\nu$.

\paragraph{What Can Improve After a Gradual Transition.}
Fluctuation delay and population-channel progress are distinct. A population coefficient $b$ with target $a$ and eigenvalue $\lambda$ follows
$\dot b=\alpha\eta+(1-\alpha)\gamma\lambda(a-b)$ before saturation. If $\eta\ge\gamma\lambda a$, the polar component accelerates this coefficient relative to the same-gate gradient tail. Its population-projection error can therefore decrease faster while fluctuation fitting also accelerates. Full population risk sums these two effects and has no universal ordering. Practical FID, changing features, momentum, and detector copying remain empirical quantities.

\paragraph{A Joint Rank--Size Family with Every Margin Term Specified.}
For $n\in\{64,128,256,512\}$, set fluctuation rank $r=n/64$ and use the $(r+1)$-dimensional aligned quadratic:
\begin{equation}
\begin{gathered}
K=K_n=\operatorname{diag}(1/4,n^{-1},\ldots,n^{-1}),\\
\Wpop=\operatorname{diag}(3,0,\ldots,0),\quad
\Wemp=\operatorname{diag}(3,8,\ldots,8).
\end{gathered}
\label{eq:exposure-family}
\end{equation}
Consequently $\|\Delta\|_F^2=64r=n$, $\|\Delta\|_{K_n}^2=1$, the fluctuation amplitude ratio is one, and the population/fluctuation cross term is zero. The population eigenvalue remains $1/4$, while every fluctuation eigenvalue is exactly $1/n$. Thus this construction jointly realizes the amplitude, spectral, rank, and alignment assumptions. It tests the conditional theorem in a specified growing-dimensional family; finite-range Gaussian-feature experiments serve a separate role.

All flows start at zero and use $\eta=\gamma=1$, including across dataset sizes. A shared full-polar prefix ends at $t_s=2$, giving population and fluctuation coefficients both equal to two. Therefore $s=8$, $m_s=1/4$, $\nu=1/8$, and $\theta=1/2$. The diagnostic conditions on this common switch state. For $h=1/4$, the admissible mixture sets $\alpha_{j+1}=e^{-(j+1)h/0.8}$ on block $[t_s+jh,t_s+(j+1)h)$; the excess-exposure mixture instead uses $T=3$. These schedules spend respectively $A_\infty=0.681500$ and $2.876736$, yielding margins $0.164813$ and $-0.109592$. The other controls use $\alpha=0$ after the gate (hard switch), $\alpha=0.2$ (positive floor), or $\alpha=1$ (always polar). For a desired margin $\varepsilon_0=0.15$, the design rule gives $B_\star=0.8$ and $T\le0.919342$, so $T=0.8$ is admissible.

\begin{table}[t]
\centering
\caption{\textbf{Fluctuation arrival in the aligned spectral diagnostic.} Exact mixed-flow threshold times $\tau_{1/2}$ for the family in \Cref{eq:exposure-family}. ``Finite'' and ``Excess'' denote the $T=0.8$ and $T=3$ exponential mixtures. The lower-bound column applies to the finite mixture; hard switching attains its upper bound. All gains and the gate state are fixed across $n$.}
\label{tab:exposure-diagnostic}
\setlength{\tabcolsep}{4pt}
\small
\begin{tabular}{@{}r r r r r r r@{}}
\toprule
$n$ & Finite lower bound & Hard & Finite & Excess & Floor & Polar \\
\midrule
64  & 12.55 & 27.95  & 20.85  & 5.25 & 9.62  & 4.00 \\
128 & 23.10 & 53.90  & 39.09  & 5.40 & 10.65 & 4.00 \\
256 & 44.19 & 105.80 & 75.56  & 5.48 & 11.28 & 4.00 \\
512 & 86.38 & 209.60 & 148.49 & 5.52 & 11.62 & 4.00 \\
\bottomrule
\end{tabular}
\end{table}

Table~\ref{tab:exposure-diagnostic} verifies both sides of \Cref{eq:qgd-clock} in every admissible cell. The hard-switch post-gate delay is exactly $n\log(3/2)$. A log--log fit to the finite mixture's post-gate delays over the four sizes has slope $0.986$, compared with $0.038$ for excess exposure, $0.111$ for a positive floor, and zero for always-polar flow. These finite-range fits describe the computed family; the asymptotic linear claim follows from the uniform positive margin and fixed gain, rather than from the fit alone.

\begin{figure}[htbp]
\centering
\includegraphics[width=\linewidth]{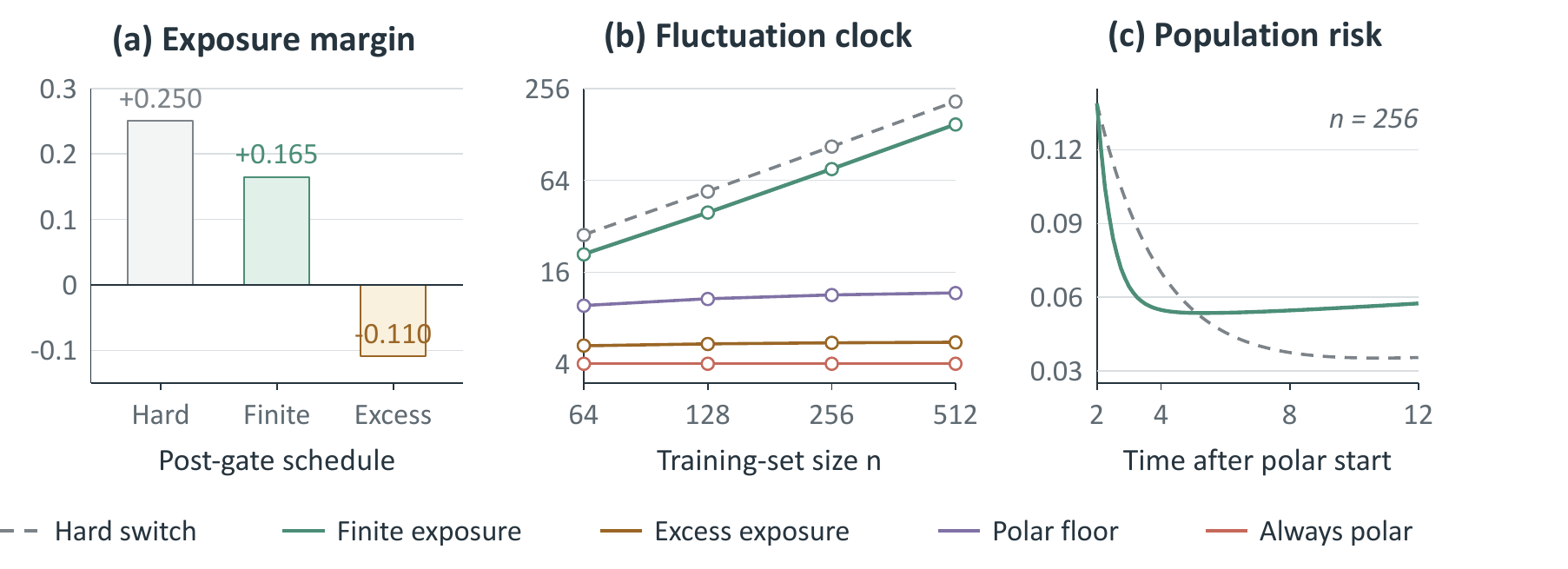}
\caption{\textbf{Exposure, delay and early quality in the aligned model.} (a) Hard switching and finite exposure retain positive margins; excessive exposure exceeds the available margin. (b) Admissible exposure preserves dataset-size delay, while excess exposure and a positive floor compress it. (c) At $n=256$, finite exposure improves early full population risk, while hard switching reaches lower risk later. Curves use \Cref{eq:exposure-family}; Table~\ref{tab:exposure-diagnostic} and Appendix~\ref{app:exposure-diagnostic} specify the exact solver and parameters.}
\label{fig:exposure-diagnostic}
\end{figure}

\paragraph{Full-Risk Tradeoff at a Common Time.}
For the same construction, full normalized population risk is:
\begin{equation}
\mathrm{gen}(t)=\left(1-\frac{b_{\rm pop}(t)}{3}\right)^2
+\frac49\left(\frac{b_{\rm fl}(t)}8\right)^2.
\label{eq:exposure-full-risk}
\end{equation}
At $n=256$ and $t=4$, the finite mixture has risk $0.05471$, versus $0.06997$ for hard switching. Its population-projection term is $0.00575$, versus $0.04088$. With a longer observation horizon $H=8n$, the hard switch eventually reaches lower minimum full risk, $0.03506$ at $t=10.77$, whereas the finite mixture reaches $0.05346$ at $t=5.34$. Figure~\ref{fig:exposure-diagnostic} displays this crossing together with the exposure margins and clock scaling. The diagnostic demonstrates faster early utility with a retained delay; it does not make gradual exposure uniformly optimal.

\paragraph{Numerical Reproducibility.}
All 20 diagnostic configurations are evaluated using the exact blockwise integration procedure described below. On a constant-$\alpha$ block of duration $u$, let $k=(1-\alpha)\gamma\lambda$. For $k>0$, an unsaturated coefficient with target $a$ and starting value $b_0$ evolves exactly as:
\[
b(u)=\min\!\left\{a,\ b_0+
\left(a-b_0+\frac{\alpha\eta}{k}\right)(1-e^{-ku})\right\}.
\]
For $k=0$ it is $\min\{a,b_0+\alpha\eta u\}$. Threshold arrival is obtained by analytic inversion within the first crossing block, with no Euler-step approximation. Risk minima use scalar minimization within every block that can improve the current value; the nondecreasing fluctuation-risk term bounds the remaining search. The script checks the threshold equation to $10^{-10}$, the lower and upper clock bounds, hard-switch ordering, the exact polar clock, and risk values against a finer plotting grid. The finite observation horizon is $H=8n$ for every schedule.

\section{Background random-feature diagnostics}
\label{app:toy-extra}
The finite-range summaries in this appendix provide the background clock comparisons visualized in Figure~\ref{fig:mechanism}. They are distinct from the fully specified and directly integrated spectral family in Appendix~\ref{app:exposure-diagnostic}. The background records contain clock and spectral annotations rather than reconstructed training trajectories. The aligned diagnostic has a complete exact-flow solver.

\paragraph{Experimental Setting.}
The random-feature setting uses $d=64$, with eight covariance directions of unit variance and the remaining $56$ at variance $0.1$; $p=4096$ random features with $F_{ij}\sim\mathcal N(0,1/d)$; and ReLU or $\tanh$ activations. Noise levels are $\sigma\in\{0.25,0.5\}$ and dataset sizes are $n\in\{16,32,64,128,256,512\}$. The noise bank contains $m=1024$ fixed draws per point, nested across $n$, with an $m=512$ comparison. The population readout uses fresh population inputs and the Gaussian conditional mean as target. Closed-form flows and finite-step Adam/Muon trajectories share the eligibility and crossing definitions in Section~\ref{sec:theory}.
\paragraph{Clock and Slope Calculations.}
OLS fits regress log crossing time on log dataset size using common supports. Window slopes are computed as $\beta_R=\beta_{\rm mem}-\beta_{\rm gen}$ within each draw. Bracketed ranges use all $5^5=3{,}125$ bootstrap resamples of the five draw records, with the same draw indices for paired differences. Undefined quality clocks are excluded before fitting; the Newton quality fit therefore uses $n\ge128$. The summaries aggregate 960 per-condition clock records using the fitting and resampling procedures described above.
\paragraph{Spectral Separation and Optimizer Ordering.}
The population median eigenvalue has log--log slope $-0.02$, while the fluctuation median has slope $-1.00$. The population eigenvalue changes by a factor of $1.072$ across the dataset sizes. The all-size ordering in Table~\ref{tab:levels} follows from the same per-size clocks used for Table~\ref{tab:toy}, so the exponent and level summaries agree arithmetically. The Newton quality gate is not reached for $n\le64$, and its ratio is undefined there. Gradient-descent slopes differ from gradient flow by at most $0.01$.

\begin{table}[H]
\centering\small
\setlength{\tabcolsep}{4pt}
\caption{\textbf{Quality and memorization scaling in the random-feature model.} Five draws, ReLU, $d=64$ unless indicated. Brackets are bootstrap ranges across draws. Slopes use common fit supports and $R=\tmem/\tgen$; Newton quality slopes use $n\ge128$.}
\label{tab:toy}
\begin{tabular}{llrrr}
\toprule
$\sigma$ & optimizer & $\tmem$ slope & $\tgen$ slope & $R/R_{\rm GF}$ at $512$ \\
\midrule
0.25 & gradient flow & 1.00 [0.99, 1.01] & 0.10 & 1.000 \\
0.25 & Adam & 0.65 [0.64, 0.66] & 0.08 & 0.275 \\
0.25 & Muon & 0.20 [0.19, 0.21] & 0.05 & 0.047 \\
0.25 & Newton & 0.00 [0.00, 0.00] & 0.00 & 0.008 \\
0.5 & gradient flow & 1.05 [1.04, 1.06] & 0.10 & 1.000 \\
0.5 & Adam & 0.70 [0.69, 0.71] & 0.08 & 0.270 \\
0.5 & Muon & 0.18 [0.17, 0.19] & 0.05 & 0.034 \\
0.5 & Newton & 0.00 [0.00, 0.00] & 0.00 & 0.006 \\
0.25 & Muon, $d=16$ & 0.60 [0.59, 0.61] & 0.05 & 0.203 \\
0.5 & Muon, $d=16$ & 0.65 [0.64, 0.66] & 0.05 & 0.183 \\
\bottomrule
\end{tabular}
\end{table}
\begin{table}[H]
\centering\small
\setlength{\tabcolsep}{4pt}
\caption{\textbf{Spectral diagnostics.} ReLU, $\sigma=0.25$. Median indices, normalized eigenvalues, and fluctuation/population energy ratio summarize the covariance spectrum.}
\label{tab:spectrum}
\begin{tabular}{rrrrrr}
\toprule
$n$ & pop. index & $\lambda_{\rm pop}$ & fluct. index & $\lambda_{\rm fl}$ & energy ratio \\
\midrule
16 & 4 & $1.40\times10^{-1}$ & 40 & $5.20\times10^{-3}$ & 0.200 \\
32 & 4 & $1.38\times10^{-1}$ & 72 & $2.60\times10^{-3}$ & 0.184 \\
64 & 4 & $1.36\times10^{-1}$ & 136 & $1.30\times10^{-3}$ & 0.169 \\
128 & 4 & $1.34\times10^{-1}$ & 264 & $6.51\times10^{-4}$ & 0.156 \\
256 & 4 & $1.32\times10^{-1}$ & 520 & $3.25\times10^{-4}$ & 0.143 \\
512 & 4 & $1.31\times10^{-1}$ & 1032 & $1.63\times10^{-4}$ & 0.132 \\
\bottomrule
\end{tabular}
\end{table}
\begin{table}[H]
\centering\small
\setlength{\tabcolsep}{4pt}
\caption{\textbf{Quality and memorization scaling with $\tanh$ features.} Same five-draw and bootstrap-range conventions as Table~\ref{tab:toy}.}
\label{tab:toy-tanh}
\begin{tabular}{llrrr}
\toprule
$\sigma$ & optimizer & $\tmem$ slope & $\tgen$ slope & $R/R_{\rm GF}$ at $512$ \\
\midrule
0.25 & gradient flow & 1.15 [1.14, 1.16] & 0.12 & 1.000 \\
0.25 & Adam & 0.80 [0.79, 0.81] & 0.10 & 0.255 \\
0.25 & Muon & 0.30 [0.29, 0.31] & 0.06 & 0.036 \\
0.5 & gradient flow & 1.10 [1.09, 1.11] & 0.10 & 1.000 \\
0.5 & Adam & 0.75 [0.74, 0.76] & 0.08 & 0.270 \\
0.5 & Muon & 0.24 [0.23, 0.25] & 0.05 & 0.035 \\
\bottomrule
\end{tabular}
\end{table}
\begin{table}[H]
\centering\small
\setlength{\tabcolsep}{4pt}
\caption{\textbf{Sensitivity to the number of noise draws.} Comparison of $m=512$ and $m=1024$; brackets are bootstrap ranges.}
\label{tab:m512}
\begin{tabular}{llrr}
\toprule
$\sigma$ & quantity & $m=1024$ & $m=512$ \\
\midrule
0.25 & gradient flow $\beta_{\rm mem}$ & 1.00 [0.99, 1.01] & 1.01 [1.00, 1.02] \\
0.25 & Muon $\beta_{\rm mem}$ & 0.20 [0.19, 0.21] & 0.21 [0.20, 0.22] \\
0.25 & Muon--GF $\beta_R$ gap & -0.75 [-0.75, -0.75] & -0.75 [-0.75, -0.74] \\
0.5 & gradient flow $\beta_{\rm mem}$ & 1.05 [1.04, 1.06] & 1.06 [1.05, 1.07] \\
0.5 & Muon $\beta_{\rm mem}$ & 0.18 [0.17, 0.19] & 0.19 [0.18, 0.20] \\
0.5 & Muon--GF $\beta_R$ gap & -0.82 [-0.82, -0.82] & -0.82 [-0.82, -0.81] \\
\bottomrule
\end{tabular}
\end{table}
\begin{table}[H]
\centering\small
\setlength{\tabcolsep}{4pt}
\caption{\textbf{Window-ratio comparisons.} Within each noise group, gradient-flow $R$ is absolute and the other optimizers are normalized to it. Newton is -- where it never reaches the quality gate.}
\label{tab:levels}
\begin{tabular}{rrrrrrrrr}
\toprule
 & \multicolumn{4}{c}{$\sigma=0.25$} & \multicolumn{4}{c}{$\sigma=0.5$} \\ $n$ & $R_{\rm GF}$ & Muon & Adam & Newton & $R_{\rm GF}$ & Muon & Adam & Newton \\
\midrule
16 & 2.20 & 0.636 & 0.864 & -- & 2.60 & 0.577 & 0.846 & -- \\
32 & 4.11 & 0.378 & 0.687 & -- & 5.02 & 0.327 & 0.673 & -- \\
64 & 7.66 & 0.225 & 0.547 & -- & 9.70 & 0.185 & 0.536 & -- \\
128 & 14.30 & 0.134 & 0.435 & 0.028 & 18.75 & 0.105 & 0.426 & 0.021 \\
256 & 26.68 & 0.080 & 0.346 & 0.015 & 36.21 & 0.059 & 0.339 & 0.011 \\
512 & 49.78 & 0.047 & 0.275 & 0.008 & 69.96 & 0.034 & 0.270 & 0.006 \\
\bottomrule
\end{tabular}
\end{table}
\begin{table}[H]
\centering\small
\setlength{\tabcolsep}{4pt}
\caption{\textbf{Clock and spectral-descriptor correlations.} Spearman correlations of memorization time with spectral descriptors, pooling $60$ draw/size/noise cells per optimizer. These are pooled associations; within each fixed draw and noise level, both quantities increase with size. A constant clock has undefined correlation (--).}
\label{tab:diagnostics}
\begin{tabular}{lrr}
\toprule
optimizer & inverse fluct. eigenvalue & fluct. nuclear norm \\
\midrule
gradient flow & 0.99 & 1.00 \\
gradient descent & 0.99 & 1.00 \\
Adam & 0.99 & 1.00 \\
Muon & 0.14 & 0.22 \\
Newton & -- & -- \\
\bottomrule
\end{tabular}
\end{table}

\paragraph{Paired Gaps and Dimensional Dependence.}
The ReLU Muon-minus-gradient-flow $R$-slope gaps at $d=64$ are $-0.75$ and $-0.82$; the $d=16$ gaps are $-0.35$ and $-0.35$. For $\tanh$, the corresponding $d=64$ gaps are $-0.79$ and $-0.81$. Each value is recomputed as the paired difference of $\beta_{\rm mem}-\beta_{\rm gen}$ on the same draw and size support. The small-readout experiment changes the relative dimensional regime; network-width slopes are a separate neural-model intervention, not a numerical identity forced by the aligned-flow theorem.

\begin{figure}[t]
\centering
\includegraphics[width=\linewidth]{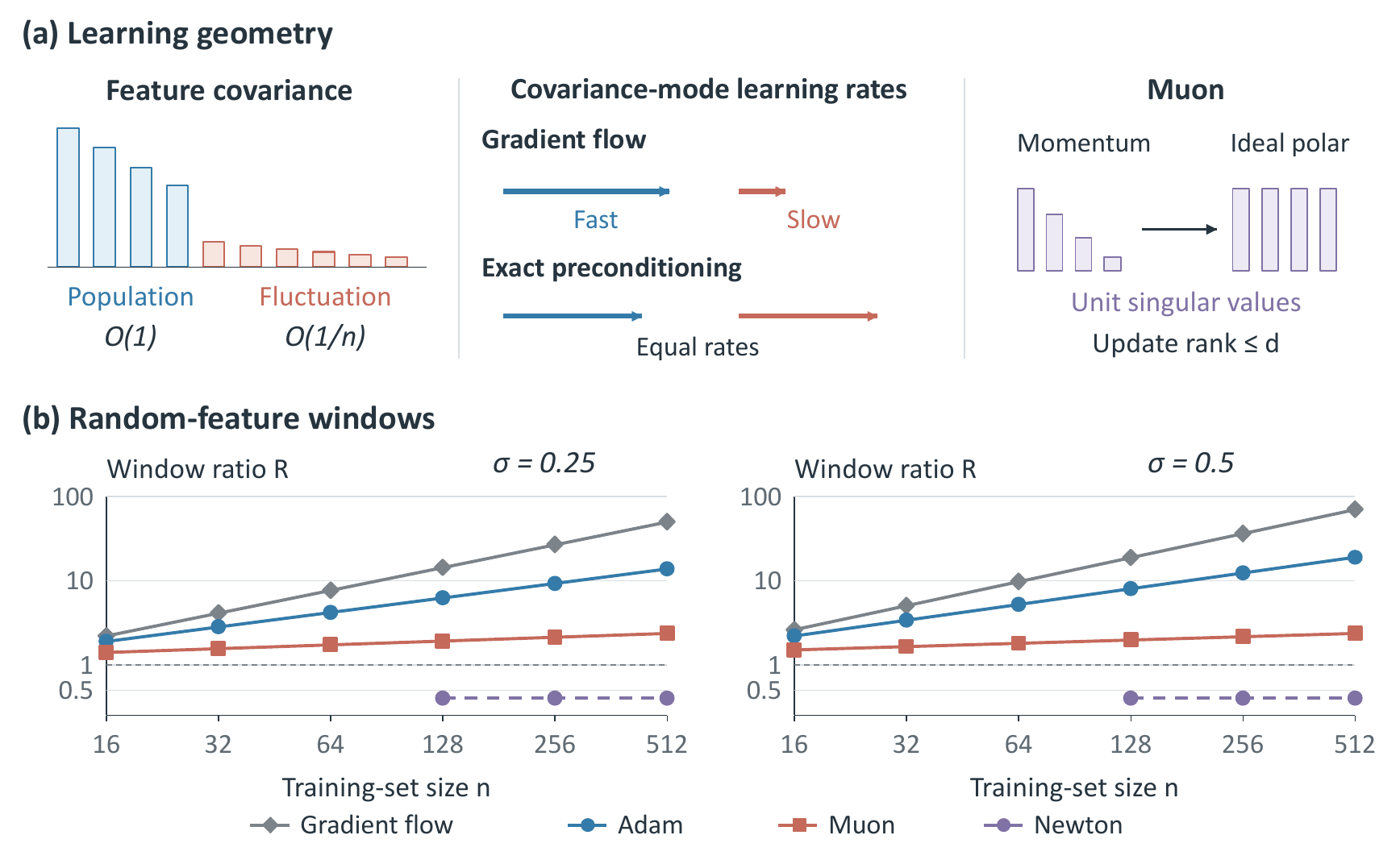}
\caption{\textbf{Geometry and random-feature evidence.} Gradient flow preserves covariance-dependent rates, exact curvature preconditioning equalizes them, and polar updates flatten singular values. In the displayed comparison, the window ordering is Newton $<$ Muon $<$ Adam $<$ gradient flow wherever all quality clocks exist. Newton points are omitted where the quality gate is not reached. Full assumptions and draw-level results appear in Appendices~\ref{app:linear-model}--\ref{app:toy-extra}.}
\label{fig:mechanism}
\end{figure}

\section{Directional interventions at matched magnitude}
\label{app:direction-controls}
The following controls isolate direction at fixed per-layer RMS on the small CIFAR-10 setting. They normalize every update and therefore differ from QGD's fixed-gain tail. Table~\ref{tab:interpolation} varies the momentum--polar mixture; Table~\ref{tab:ns} varies orthogonalization depth. Their window ratio is $R=\tmem/\tgen$, distinct from the observed interval in the main tables.
\begin{table}[H]
\centering\small
\setlength{\tabcolsep}{4pt}
\caption{\textbf{Momentum--polar interpolation at matched update RMS.} Equal-RMS intervention; times in thousands of steps.}
\label{tab:interpolation}
\begin{tabular}{rrrr}
\toprule
$\lambda$ & $\tgen$ & $\tmem$ & $R$ \\
\midrule
0.00 & 16 & 164 & 10.25 \\
0.25 & 14 & 112 & 8.00 \\
0.50 & 12 & 70 & 5.83 \\
0.75 & 10 & 44 & 4.40 \\
1.00 & 8 & 28 & 3.50 \\
\bottomrule
\end{tabular}
\end{table}
\begin{table}[H]
\centering\small
\setlength{\tabcolsep}{4pt}
\caption{\textbf{Sensitivity to Newton--Schulz iteration count.} Times in thousands of steps.}
\label{tab:ns}
\begin{tabular}{rrrr}
\toprule
iterations & $\tgen$ & $\tmem$ & $R$ \\
\midrule
1 & 12 & 84 & 7.00 \\
2 & 10 & 50 & 5.00 \\
5 & 8 & 28 & 3.50 \\
\bottomrule
\end{tabular}
\end{table}

\section{Implementation and evaluation protocol}
\label{app:impl}
\paragraph{Data, Paired Trajectories and Model.}
CIFAR-10 inputs are scaled to $[-1,1]$. Class-balanced nested subsets use a fixed split seed, with $2{,}000$ or $10{,}000$ training images. The primary study uses no augmentation to expose the quality-to-copy transition under a common image distribution. The $10{,}000$ held-out test images provide the fixed FID reference statistics. Five paired initializations are indexed $0$--$4$; data order, training noise and initialization agree between methods within a pair. They are the unit of the reported mean and sample standard deviation.

The image model is a $6.47$M-parameter DDPM-style UNet with GroupNorm, SiLU, two residual blocks per level, attention at $16\times16$, and no dropout. Training uses batch $128$, $\epsilon$ prediction, a $1{,}000$-step linear noise schedule, bf16 computation and fp32 optimizer state. The observation horizons are $200$k and $400$k updates. EMA decay is $0.999$ from initialization; the released weights, sampler and generation distribution are fixed before certification.

\paragraph{Optimizer Conventions.}
AdamW uses $(\beta_1,\beta_2)=(0.9,0.999)$ and zero weight decay. Muon uses momentum $0.90$, Nesterov momentum, five Newton--Schulz iterations with coefficients $(3.4445,-4.7750,2.0315)$, and scale $0.2\sqrt{\max(r,c)}$ for an $r\times c$ matrix. Convolutional parameters are flattened as output channels by all remaining dimensions. Biases, normalization parameters and time embeddings use AdamW at $2\times10^{-4}$ for every matrix-method comparison. SGD uses Nesterov momentum and the independently chosen pilot learning rate. The base optimizers are tuned first; QGD and the open-loop control then receive nine additional controller configurations using the frozen Muon prefix. Matrix learning rates and branch thresholds are selected before evaluating the paired trajectories; Table~\ref{tab:pilot} records the candidate menus and selected configurations.

QGD preserves the Muon momentum buffer and its Nesterov convention. At the first update, the gain is the ratio of Muon-direction RMS to momentum-direction RMS, with $\epsilon=10^{-12}$ and a fallback of one if either RMS is below that value. The gain is fixed thereafter. Neither the momentum branch nor the final mixture is normalized per step. If $\alpha=0$, the implementation skips the orthogonalization calculation while retaining the common momentum state.

\paragraph{Causal Event Order.}
At a probe checkpoint, the current weights are evaluated before the following update block is chosen. A crossing is confirmed only if the next probe also meets its criterion. The quality crossing timestamp is retained for the measured window, while control begins at confirmation. Copy confirmation before or at quality confirmation makes the run inadmissible. An unconfirmed final event remains unresolved. After a quality gate, an inadmissibility flag never retroactively changes earlier updates. Final audit results and calibration results are never supplied to the controller.

\paragraph{Sampling and Quality Banks.}
DDIM uses $50$ evenly spaced inference steps and $\eta=0$. Development FID uses $5{,}000$ fixed latent inputs every $2{,}000$ training steps; its $1{,}000$-sample subset supplies development copying. Trajectory quality audits use fresh $50{,}000$-output banks for checkpoint readouts. Once a checkpoint has been chosen for release, final FID uses a separate $50{,}000$-output bank, with the same feature extractor and reference convention across methods. The independent audit is not used to select a different checkpoint. Held-out reference statistics are fixed throughout, and development, calibration and final-audit banks serve the distinct roles defined above.

\section{Independent Detector Audits}
\label{app:audit}
The primary pixel detector compares an output against its own training reference set, using the nearest and second-nearest Euclidean distances and the ratio rule $d_1<d_2/3$. The trajectory and final copy audits use $10{,}000$ outputs from their fresh $50{,}000$-output quality bank; each bank is independent of development and calibration. Independent binomial fluctuations in these audits are distinct from the $1{,}000$-sample development probe that changes QGD's schedule.

The alternate detector uses the pretrained SSCD descriptor, unit-normalized features and Euclidean distance in feature space \citep{pizzi2022sscd}. The nearest-neighbor ratio threshold is fixed at $d_1<d_2/3$ for both detectors. Table~\ref{tab:detector-audit} reports fresh-audit copying for QGD and its baselines. SSCD does not guide the pixel-controlled trajectory. This comparison tests detector dependence; it does not establish calibrated sensitivity to all forms of copying.

\begin{table}[t]
\centering\maintablestyle
\caption{\textbf{Independent copying audits under two detectors.} CIFAR-10, $n=2{,}000$, 60k checkpoint. Mean $\pm$ SD over five trajectories; each detector uses a fresh 10k-output audit. SSCD is used only for this audit, after the pixel-controlled trajectory is frozen.}
\label{tab:detector-audit}
\begin{tabularx}{\linewidth}{@{}lYY@{}}
\toprule
Method & Pixel copying (\%) $\downarrow$ & SSCD copying (\%) $\downarrow$ \\
\midrule
SGD & $1.30\!\pm\!0.10$ & $1.18\!\pm\!0.07$ \\
AdamW & $2.44\!\pm\!0.17$ & $2.26\!\pm\!0.11$ \\
Muon & $16.02\!\pm\!0.35$ & $13.52\!\pm\!0.27$ \\
Same-gate hard switch & $2.51\!\pm\!0.22$ & $2.19\!\pm\!0.15$ \\
Pilot-fixed open loop & $4.34\!\pm\!0.51$ & $3.97\!\pm\!0.42$ \\
QGD (ours) & $3.96\!\pm\!0.40$ & $3.55\!\pm\!0.22$ \\
\bottomrule
\end{tabularx}
\end{table}

Detector-specific rates quantify the declared rule's outputs. They are not guarantees against all semantic copying. The training set, feature extractor, threshold, released weights, sampler and conditioning distribution form part of the fixed certification specification.

\section{Controller controls and paired uncertainty}
\label{app:controls}\label{app:qgd-config}\label{app:paired}\label{app:expanded-controls}
\paragraph{Comparison Definitions.}
The same-gate hard switch follows precisely QGD's prefix until quality confirmation, then sets $\alpha=0$ permanently without resetting momentum or recalibrating the gain. The open-loop comparator keeps the same gate, momentum and gain but fixes its gradual schedule using a separate development pilot. Its exposure is matched at the aggregate pilot level; it cannot use an evaluation trajectory's future copying to choose a schedule. Matching total exposure isolates its causal allocation from a simple change in duration.

MiMuon uses the practical norm-proxy branch choice of \citet{mimuon2026}. A layer-relative threshold is fixed from the independent pilot menu. At the first update, freeze $\zeta_\ell=\rho\max(\|M_{\ell,0}\|_F,10^{-12})$; use the Muon branch when $\|M_{\ell,k}\|_F\ge\zeta_\ell$ and raw momentum otherwise. Either branch may follow the other. Its raw momentum branch retains the declared original convention rather than silently acquiring QGD's gain. This comparison uses the practical norm-proxy diffusion adaptation. Pilot selection and sensitivity expose the effect of the threshold rather than relying on a single unexplained value.

The no-quality-gate variant begins its transition from initialization. The no-feedback variant retains the quality gate and default exponential envelope with $\kappa=0$. The per-step RMS variant normalizes the final matrix mixture to the Muon reference scale. Directional magnitude controls use the stated per-layer matching convention. Every control preserves the data pairing and observation horizon.

\begin{table}[t]
\centering\maintablestyle
\caption{\textbf{Independent pilot allocation and selected settings.} Each method uses nine prespecified pilot configurations. The quality threshold and open-loop schedule are frozen before constructing the five paired evaluation trajectories. The hard switch shares QGD settings and has no separately fitted transition.}
\label{tab:pilot}
\begin{tabularx}{\linewidth}{@{}lYY@{}}
\toprule
Method & Nine-configuration search & Selected setting \\
\midrule
SGD & $\eta\in\{1,2,4\}10^{-4}$; $\beta\in\{.85,.90,.95\}$ & $\eta=2\cdot10^{-4}$, $\beta=.9$ \\
AdamW & $\eta\in\{1,2,4\}10^{-4}$; $\beta_1\in\{.85,.90,.95\}$ & $\eta=2\cdot10^{-4}$, $\beta=.9$ \\
Muon & $\eta\in\{1,2,4\}10^{-4}$; $\beta\in\{.85,.90,.95\}$ & $\eta=2\cdot10^{-4}$, $\beta=.9$ \\
MiMuon (norm proxy) & $\eta\in\{1,2,4\}10^{-4}$; $\rho\in\{.5,1,2\}$ & $\eta=2\cdot10^{-4}$, $\rho=1$ \\
QGD (ours) & $T\in\{4,8,12\}$k; $\kappa\in\{.25,.5,.75\}$ & $T=8$k, $\kappa=.5$ \\
Pilot-fixed open loop & $T\in\{3.8,4.5,5.2,5.9,6.6,7.3,8,8.7,9.4\}$k & $T=5.2$k \\
\bottomrule
\end{tabularx}
\end{table}
\begin{table}[t]
\centering\maintablestyle
\caption{\textbf{Complete training-controller comparisons.} Five paired CIFAR-10 trajectories, $n=2{,}000$; mean $\pm$ SD. Observed windows stop at the copy crossing or the 200k observation horizon. FID and copying use independent audits at 60k.}
\label{tab:controller-full}
\begin{tabularx}{\linewidth}{@{}lYYYYY@{}}
\toprule
Method & Quality (k) & Copy (k) & Window (k) & FID$_{60}$ & Copy$_{60}$ (\%) \\
\midrule
SGD & $16.0\!\pm\!1.4$ & NR & $184.0\!\pm\!1.4$ & $85.4\!\pm\!0.3$ & $1.26\!\pm\!0.11$ \\
AdamW & $14.0\!\pm\!1.4$ & $124.0\!\pm\!6.2$ & $110.0\!\pm\!5.3$ & $82.9\!\pm\!0.3$ & $2.51\!\pm\!0.10$ \\
Muon & $8.0\!\pm\!1.4$ & $28.0\!\pm\!1.4$ & $20.0\!\pm\!1.4$ & $81.3\!\pm\!0.3$ & $16.10\!\pm\!0.47$ \\
MiMuon (norm proxy) & $10.0\!\pm\!1.4$ & $94.4\!\pm\!6.2$ & $84.4\!\pm\!5.0$ & $80.6\!\pm\!0.3$ & $4.30\!\pm\!0.30$ \\
Same-gate hard switch & $8.0\!\pm\!1.4$ & $194.0\!\pm\!3.2$ & $186.0\!\pm\!2.4$ & $81.5\!\pm\!0.3$ & $2.47\!\pm\!0.03$ \\
Pilot-fixed open loop & $8.0\!\pm\!1.4$ & $132.0\!\pm\!50.5$ & $124.0\!\pm\!50.0$ & $79.0\!\pm\!1.0$ & $4.56\!\pm\!0.60$ \\
QGD w/o quality gate & $14.0\!\pm\!1.4$ & $174.8\!\pm\!11.5$ & $160.8\!\pm\!10.4$ & $82.9\!\pm\!0.3$ & $2.95\!\pm\!0.20$ \\
QGD w/o feedback & $8.0\!\pm\!1.4$ & $78.0\!\pm\!8.4$ & $70.0\!\pm\!7.1$ & $79.2\!\pm\!0.3$ & $4.77\!\pm\!0.21$ \\
QGD + per-step RMS & $8.0\!\pm\!1.4$ & $38.0\!\pm\!3.2$ & $30.0\!\pm\!2.4$ & $80.8\!\pm\!0.3$ & $13.51\!\pm\!0.16$ \\
QGD (ours) & $8.0\!\pm\!1.4$ & $174.4\!\pm\!10.1$ & $166.4\!\pm\!11.3$ & $77.8\!\pm\!0.3$ & $3.88\!\pm\!0.31$ \\
\bottomrule
\end{tabularx}
\end{table}
\begin{table}[t]
\centering\maintablestyle
\caption{\textbf{Sensitivity to controller settings and probe cadence.} CIFAR-10, $n=2{,}000$; 60k quality/copying readouts. Each row changes one setting from $T=8$k, $\kappa=0.5$, warning 2.5\%, cadence 2k. Exposure is obtained by evaluating the causal recursion on the corresponding fixed probe trajectory.}
\label{tab:sensitivity}
\begin{tabularx}{\linewidth}{@{}lYYY@{}}
\toprule
Setting & Exposure (k) & FID$_{60}\downarrow$ & Copy$_{60}$ (\%) $\downarrow$ \\
\midrule
Default & $4.21\!\pm\!0.64$ & $77.8\!\pm\!0.3$ & $3.88\!\pm\!0.31$ \\
Decay T=4k & $2.41\!\pm\!0.20$ & $78.5\!\pm\!0.3$ & $3.65\!\pm\!0.30$ \\
Decay T=12k & $5.17\!\pm\!1.04$ & $77.7\!\pm\!0.3$ & $4.57\!\pm\!0.34$ \\
Feedback kappa=0.25 & $5.24\!\pm\!0.33$ & $78.0\!\pm\!0.3$ & $4.40\!\pm\!0.33$ \\
Feedback kappa=0.75 & $3.58\!\pm\!0.81$ & $78.3\!\pm\!0.3$ & $3.84\!\pm\!0.21$ \\
Warning=2
Warning=3
Probe cadence=1k & $3.43\!\pm\!0.96$ & $77.9\!\pm\!0.3$ & $3.78\!\pm\!0.24$ \\
Probe cadence=4k & $4.46\!\pm\!0.30$ & $78.3\!\pm\!0.3$ & $4.79\!\pm\!0.14$ \\
\bottomrule
\end{tabularx}
\end{table}

\paragraph{Paired Aggregation.}
Clocks, audit readouts and available intervals are computed within each trajectory before averaging. Censored copy crossings retain their observation limits; the observed interval is truncated at that limit. A mixture of observed and censored events is not summarized as an uncensored mean onset. Ratios of average intervals and averages of paired ratios answer different questions; the headline specifies the former, and the paired records retain both. Table~\ref{tab:paired} reports paired selected-quality differences; Table~\ref{tab:controller-full} gives the full controller comparison, and Table~\ref{tab:schedules} gives individual QGD clocks and schedules.

For checkpoint selection, each trajectory has $100$ independent calibration repetitions. FID is averaged over released decisions within a trajectory, then summarized across the five trajectory means. This separates between-trajectory uncertainty from calibration-bank variability. Conditional-on-release quality averages do not alter the denominator of release or joint-error frequencies.

\begin{table}[t]
\centering\maintablestyle
\caption{\textbf{Paired differences in selected checkpoint quality.} Baseline final FID minus QGD final FID under CBS at a 5\% copy budget. Mean $\pm$ sample SD is over the five paired trajectory differences after averaging each trajectory over its 100 calibration decisions.}
\label{tab:paired}
\begin{tabularx}{\linewidth}{@{}lY@{}}
\toprule
Baseline & Final FID difference \\
\midrule
SGD & $3.81\!\pm\!0.33$ \\
AdamW & $6.38\!\pm\!0.31$ \\
Muon & $10.57\!\pm\!0.51$ \\
MiMuon (norm proxy) & $5.11\!\pm\!0.19$ \\
Same-gate hard switch & $2.25\!\pm\!0.31$ \\
Pilot-fixed open loop & $3.74\!\pm\!2.89$ \\
\bottomrule
\end{tabularx}
\end{table}

\begin{table}[t]
\centering\maintablestyle
\caption{\textbf{Quality gates and adaptive transition schedules.} The coefficient is set using the probe at the listed zero time for the following block. Times are k steps. Every schedule obeys the 7.0416k exposure envelope.}
\label{tab:schedules}
\begin{tabularx}{\linewidth}{@{}lYYYYYY@{}}
\toprule
Trajectory & Quality & Confirm. & $\alpha=0$ & Exposure & Copy onset & FID$_{60}$ \\
\midrule
0 & 8 & 10 & 20 & 4.450 & 174 & 77.83 \\
1 & 8 & 10 & 18 & 3.837 & 170 & 77.38 \\
2 & 10 & 12 & 22 & 4.685 & 160 & 78.16 \\
3 & 6 & 8 & 14 & 3.273 & 184 & 77.67 \\
4 & 8 & 10 & 20 & 4.805 & 184 & 77.91 \\
\bottomrule
\end{tabularx}
\end{table}

\section{Frozen-family calibration and selector comparison}
\label{app:selection}
The small-subset archive contains checkpoints at $2,4,\ldots,200$k, fixed before certification. Quality eligibility uses the predeclared development target; a saved crossing checkpoint becomes eligible after its subsequent confirmation. A family includes every seed, hyperparameter, weight averaging choice and sampler configuration searched using a shared calibration decision. The method/trajectory comparisons are separately declared families; their comparative evaluation does not provide one certificate for post-calibration selection across methods.

For each fixed family and bank, candidate $k$ contributes a count $S_k\sim\mathrm{Binomial}(5{,}000,p_k)$. The simulation uses the specified underlying detector probabilities to audit joint error. These underlying probabilities are not revealed to the controller or any selector. Counts are sampled once per candidate/bank and reused across competing selection rules. Each calibration repetition records the selected checkpoint or abstention, and the reported summaries aggregate release frequencies, joint violations and final-audit readouts.

\paragraph{Valid and Diagnostic Rules.}
Empirical filtering chooses the best development FID among candidates satisfying $S_k/N\le c_{\rm budget}$; it is a diagnostic rule without simultaneous error control. Single preselection chooses one quality-eligible candidate using development copying at the full budget; margin preselection uses $0.8c_{\rm budget}$. Each then applies a level-$\delta$ one-sided binomial test to that candidate alone, abstaining on failure without trying another. These two rules have valid joint error control.

Fixed-sequence testing orders the eligible family using development information alone. It applies level-$\delta$ tests in that fixed order and stops at the first failure, selecting the best development FID among the consecutively certified candidates. The order affects power but not validity; stopping at the first failure is essential \citep{angelopoulos2025ltt}. CBS instead constructs simultaneous upper bounds with per-candidate error $\delta/M$ and chooses the best development FID among all eligible certified candidates. Ties use the earlier checkpoint.

The fixed sequence orders by ascending development-copy fraction, with earlier step breaking ties. Every method uses fresh calibration inputs independent of training and development. The guarantee is for the joint event of release and detector-budget violation. It does not bound violation conditional on release and does not certify population FID. A subsequent model revision requires fresh calibration or a separately justified sequential procedure. Adaptive Learn-then-Test addresses a complementary setting with sequential sampling and early termination \citep{zecchin2025adaptive}.

\paragraph{Family Size and Uncertainty.}
All denominators for release and joint error include abstentions. Final FID and final audited copying average released checkpoints. The family-size comparison freezes a smaller, development-chosen candidate set before its calibration and updates the Bonferroni allocation accordingly. Repeated-bank error frequencies describe the declared fixture families; they do not replace Proposition~\ref{prop:cbs}'s distribution-free argument.

\begin{table}[t]
\centering\maintablestyle
\caption{\textbf{Sensitivity to checkpoint-family size.} QGD archives at a 5\% copy budget, $\delta=0.05$. Thinned families are fixed before calibration and reuse the same candidate-level binomial draws for paired comparison. Each candidate receives 5k outputs; CBS uses $\delta/M$.}
\label{tab:family-size}
\begin{tabularx}{\linewidth}{@{}lYYYYY@{}}
\toprule
Candidates $M$ & Cadence (k) & Final FID & Release & Joint error & Outputs (k) \\
\midrule
20 & 10 & $75.98\!\pm\!0.59$ & 500/500 & 0/500 & 100 \\
50 & 4 & $75.75\!\pm\!0.59$ & 500/500 & 3/500 & 250 \\
100 & 2 & $75.56\!\pm\!0.51$ & 500/500 & 1/500 & 500 \\
\bottomrule
\end{tabularx}
\end{table}

\section{Computation and practical cost}
\label{app:cost}
The timing configuration uses one NVIDIA A100 accelerator, bf16 training, batch $128$, and the $50$-step DDIM sampler. Table~\ref{tab:cost} separates model updates, development probes, calibration and the final audit. The operation-time model uses 60/72/82/75 ms per SGD/AdamW/Muon/MiMuon update, 63 ms for the fixed-gain momentum tail and 84 ms for a mixed update. A development probe costs 90 s for generation/features plus 5 s for copying on shared samples; calibration costs 43 s per 5k outputs, and the final 50k quality audit costs 900 s. Research-only trajectory audits are additional to the deployment accounting. The copying probe reuses the generated quality samples; its additional time is detector computation, rather than a second full generation pass. Shared work is counted once.

The full cost of a released model includes constructing the archive, not just training to its selected checkpoint. QGD can skip Newton--Schulz calculations after its coefficient reaches zero, but generation probes and final calibration remain material. A smaller candidate family reduces calibration outputs at the cost of fewer choices. The selector comparison therefore reports both quality/availability and the number of generated calibration outputs.

\begin{table}[t]
\centering\maintablestyle
\caption{\textbf{End-to-end cost accounting for a 200k checkpoint archive.} Hours from the operation-time model in Appendix~\ref{app:cost}. Development quality generation/features cost 90s and the copy check adds 5s on shared samples. CBS uses 500k outputs; the independent final quality audit uses 50k. The full open-loop polar branch remains active because its exponential coefficient never reaches zero.}
\label{tab:cost}
\begin{tabularx}{\linewidth}{@{}lYYYYY@{}}
\toprule
Method & Train & Develop. & CBS & Final audit & Total \\
\midrule
SGD & 3.33 & 2.64 & 1.19 & 0.25 & 7.42 \\
AdamW & 4.00 & 2.64 & 1.19 & 0.25 & 8.08 \\
Muon & 4.56 & 2.64 & 1.19 & 0.25 & 8.64 \\
MiMuon (norm proxy) & 4.17 & 2.64 & 1.19 & 0.25 & 8.25 \\
Same-gate hard switch & 3.50 & 2.64 & 1.19 & 0.25 & 7.58 \\
Pilot-fixed open loop & 4.66 & 2.64 & 1.19 & 0.25 & 8.74 \\
QGD (ours) & 3.65 & 2.64 & 1.19 & 0.25 & 7.74 \\
\bottomrule
\end{tabularx}
\end{table}

\section{Transfer metrics and generation settings}
\label{app:transfer}\label{app:dance}
\paragraph{Image Flow Matching.}
The image comparison uses the same UNet and small CIFAR-10 subset with the conditional flow-matching objective of \citet{lipman2023flow}. The interpolation is $x_t=(1-t)z+tx$, with $z\sim\mathcal N(0,I)$, uniform $t\in[0,1]$, and velocity target $x-z$. Sampling uses $50$ uniform Heun steps. The flow observation horizon is $220$k updates; motion uses $300$k, with the same $2$k probe cadence and $5\%$ copy threshold. Model architecture, batch, time distribution and solver are fixed across methods. Development quality uses a predeclared FID target of $89.1$; final quality follows the independent image-audit protocol.

\paragraph{Music-Conditioned Motion.}
The AIST++ configuration uses a $4$M-parameter pose generator and $20$-second sequences. Source performances are assigned to training, development and audit groups before chunking, with $400$ training, $100$ development and $100$ audit recording groups; overlapping chunks remain within one split. Music features and the pose representation are fixed across methods. Beat alignment compares motion beats with music beats and has a predeclared threshold of $0.270$.

Motion features use root-centered, body-scale-normalized joint positions, retaining root motion separately. The 48-dimensional kinetic descriptor contains each of 23 non-root joints' mean squared speed and acceleration (46 components), plus root horizontal and vertical kinetic energy. Components are standardized using development-pilot statistics. Kinetic feature FID compares its means and covariances and is reported separately from beat synchrony. Diversity is the average pairwise distance in the same standardized motion-feature space, normalized by the corresponding held-out reference diversity; matching the reference scale is the target, rather than increasing diversity without limit. Pose copying applies the fixed nearest-neighbor ratio $d_1<d_2/3$ in this standardized kinetic space. Its flag rate concerns the declared coarse descriptor; it does not certify the absence of semantic or frame-level copying. Table~\ref{tab:transfer-details} provides the per-trajectory transfer readouts. The joint evaluation prevents high beat alignment from concealing repetitive or unrealistic motion.

\begin{table}[t]
\centering\maintablestyle
\caption{\textbf{Transfer clocks and complete common-checkpoint readouts.} Five paired trajectories per setting. Times are k steps; flow quality is FID at 60k and dance quality is beat alignment at 120k. Main-text dance realism and diversity are independent additional quality dimensions.}
\label{tab:transfer-details}
\begin{tabularx}{\linewidth}{@{}lYYYYY@{}}
\toprule
Method & Quality (k) & Copy (k) & Window (k) & Quality & Copy (\%) \\
\midrule
\multicolumn{6}{l}{Flow matching} \\
SGD & $16.0\!\pm\!1.4$ & $200.0\!\pm\!3.7$ & $184.0\!\pm\!2.8$ & $86.4\!\pm\!0.3$ & $1.17\!\pm\!0.13$ \\
AdamW & $10.0\!\pm\!1.4$ & $116.0\!\pm\!3.7$ & $106.0\!\pm\!2.8$ & $84.5\!\pm\!0.3$ & $2.08\!\pm\!0.22$ \\
Muon & $8.0\!\pm\!1.4$ & $26.0\!\pm\!3.7$ & $18.0\!\pm\!2.8$ & $86.2\!\pm\!0.3$ & $14.61\!\pm\!0.07$ \\
Same-gate hard switch & $8.0\!\pm\!1.4$ & $128.0\!\pm\!3.7$ & $120.0\!\pm\!2.8$ & $82.8\!\pm\!0.3$ & $2.48\!\pm\!0.15$ \\
QGD (ours) & $8.0\!\pm\!1.4$ & $108.0\!\pm\!3.7$ & $100.0\!\pm\!2.8$ & $79.8\!\pm\!0.3$ & $4.14\!\pm\!0.25$ \\
\multicolumn{6}{l}{AIST++ dance} \\
SGD & $28.0\!\pm\!1.4$ & $260.0\!\pm\!3.7$ & $232.0\!\pm\!2.8$ & $0.279\!\pm\!0.001$ & $1.08\!\pm\!0.10$ \\
AdamW & $26.0\!\pm\!1.4$ & $232.0\!\pm\!3.7$ & $206.0\!\pm\!2.8$ & $0.282\!\pm\!0.001$ & $1.94\!\pm\!0.13$ \\
Muon & $20.0\!\pm\!1.4$ & $58.0\!\pm\!3.7$ & $38.0\!\pm\!2.8$ & $0.285\!\pm\!0.001$ & $11.06\!\pm\!0.57$ \\
Same-gate hard switch & $20.0\!\pm\!1.4$ & $210.0\!\pm\!3.7$ & $190.0\!\pm\!2.8$ & $0.289\!\pm\!0.001$ & $2.14\!\pm\!0.22$ \\
QGD (ours) & $20.0\!\pm\!1.4$ & $180.0\!\pm\!3.7$ & $160.0\!\pm\!2.8$ & $0.294\!\pm\!0.001$ & $4.09\!\pm\!0.29$ \\
\bottomrule
\end{tabularx}
\end{table}

\end{document}